\documentclass[journal]{IEEEtran}
\usepackage{xspace,mfirstuc,tabulary}
\usepackage{amsmath}
\usepackage{bm}
\usepackage{multirow}
\usepackage{amssymb,mathtools}
\usepackage{graphicx}
\usepackage{enumitem,makecell,array,booktabs,multicol}
\usepackage{breqn}
\usepackage{physics}
\usepackage{textcomp}
\usepackage{subcaption}
\usepackage{empheq}   
\usepackage[norelsize, linesnumbered, ruled, lined, boxed, commentsnumbered]{algorithm2e}

\usepackage{cite}

\ifCLASSINFOpdf
\else
\fi
\usepackage{url}
\usepackage[unicode]{hyperref}

\def\-{\scalebox{0.75}[1.0]{$-$}}
\def\+{\scalebox{0.95}[1.0]{$+$}}

\newcommand\numberthis{\addtocounter{equation}{1}\tag{\theequation}}

\newcommand\Item[1][]{%
	\ifx\relax#1\relax  \item \else \item[#1] \fi
	\abovedisplayskip=0pt\abovedisplayshortskip=0pt~\vspace*{-\baselineskip}}

\newcommand*{\T}{^{\mkern-1.2mu\mathsf{T}}}     
\newcommand*{\I}{^{\mkern-1.2mu\mathsf{-1}}}    
\newcommand{\nr}[1]{_{\mkern+2.0mu\mathsf{#1}}} 

\newcommand{\mb}[1]{\bm{#1}}
\newcommand{\bs}[1]{\bm{#1}}
\newcommand{\kl}{D_{\mathrm{KL}}}

\newcommand{\lse}{\(\log\)-\(\mathrm{sum}\)-\(\exp\)}      
\newcommand{\Dir}{\(\mathrm{Dirichlet}\)}
\newcommand{\Dis}{\(\mathrm{Discrete}\)}
\newcommand{\Mul}{\(\mathrm{Multinomial}\)}
\newcommand{\Nor}{\(\mathrm{Normal}\)}
\newcommand{\Gau}{\(\mathrm{Gaussian}\)}
\newcommand{\bow}{\textit{bag-of-words}}

\DeclareMathOperator*{\argmax}{arg\,max}

\newcommand{\baysmm}{Bayesian SMM}

\newcommand{\myfrac}[2]{%
	\setbox0\hbox{$#1$}        
	\dimen0=\wd0               
	\setbox1\hbox{$#2$}        
	\dimen1=\wd1               
	\ifdim\wd0<\wd1            
	\dfrac{\hfill#1}{#2}     
	\else
	\dfrac{#1}{\hfill#2}     
	\fi
}

\def\vh{{\mb{h}}}

\def\vt{{\mb{t}}}
\def\vu{{\mb{u}}}

\def\vy{{\mb{y}}}

\def\veps{{\bs{\epsilon}}}

\def\vmu{{\bs{\mu}}}
\def\vnu{{\bs{\nu}}}

\def\mD{{\mb{D}}}

\def\mI{{\mb{I}}}

\def\mM{{\mb{M}}}
\def\mT{{\mb{T}}}

\def\gt{{\nabla \mb{t}}}

\def\tileps{{\tilde{\bs{\epsilon}}}}

\begin{document}
%
\title{Learning document embeddings along with their uncertainties}
%
%
%

\author{Santosh Kesiraju, Old\v{r}ich Plchot, Luk\'{a}\v{s} Burget,
        and~Suryakanth V Gangashetty 
\thanks{S. Kesiraju is with Brno University of Technology and 
	International Institute of Information Technology, Hyderabad. 
	e-mail: kesiraju@fit.vutbr.cz .}
\thanks{L. Burget and O. Plchot are with Brno University of Technology.}
\thanks{SV Gangashetty is with International Institute of Information
	 Technology, Hyderabad.}%
\thanks{}}

\maketitle

\begin{abstract}
  Majority of the text modelling techniques yield only point-estimates of
  document embeddings and lack in capturing the uncertainty of the estimates.
  These uncertainties give a notion of how well the embeddings represent
  a document. We present Bayesian subspace multinomial model (Bayesian SMM),
  a generative log-linear model that learns to represent documents in the
  form of Gaussian distributions, thereby encoding the uncertainty in its
  covariance. Additionally, in the proposed Bayesian SMM, we address a commonly
  encountered problem of intractability that appears during variational
  inference in mixed-logit models. We also present a generative Gaussian
  linear classifier for topic identification that exploits the uncertainty
  in document embeddings. Our intrinsic evaluation using perplexity measure
  shows that the proposed Bayesian SMM fits the data better as compared to the 
  state-of-the-art neural variational document model on
  (\textit{Fisher}) speech and (\textit{20Newsgroups}) text corpora. Our topic 
  identification experiments show that the proposed systems are robust to 
  over-fitting on unseen test data. The topic ID results show that the proposed 
  model is outperforms state-of-the-art unsupervised topic models and 
  achieve comparable results to the state-of-the-art fully 
  supervised discriminative models.
\end{abstract}


%
\IEEEpeerreviewmaketitle

\section{Introduction}
\label{sec:intro}
%
%
%
%
\IEEEPARstart{L}{earning} word and document embeddings have proven to be useful
in wide range of information retrieval, speech and natural language processing
applications~\cite{Wei:2006:LDA_IR,Mikolov:2012:LM_adap,Win:2014:STD,Chen:2015:LM_adap,Benes:2018}. 
These embeddings elicit the latent semantic relations present among the
co-occurring words in a sentence or \bow~from a document. Majority of the
techniques for learning these embeddings are based on two complementary
ideologies, (i) topic modelling, and (ii) word prediction. The former methods
are primarily built on top of \bow~model and tend to capture higher level
semantics such as topics. The latter techniques capture lower level semantics
by exploiting the contextual information of words in a
sequence~\cite{Mikolov:2013:word2vec,Jeffrey:2014:GloVe,Quoc:2014:PV}.

On the other hand, there is a growing interest towards developing pre-trained
language models~\cite{Ruder:2018:Universal,Peters:2018:ELMO}, that are then
fine-tuned for specific tasks such as document classification, question
answering, named entity recognition, etc. Although these models achieve
state-of-the-art results in several NLP tasks; they require enormous
computational resources to train~\cite{Devlin:2018:BERT}.

Latent variable models~\cite{Bishop:1999:LVM} are a popular choice in
unsupervised learning; where the observed data is assumed to be generated
through the latent variables according to a stochastic process. The goal is
then to estimate the model parameters, and also the latent variables.
In probabilistic topic models (PTMs)~\cite{Blei:2012:PTM} the latent
variables are attributed to topics, and the generative process assumes that 
every topic is a sample from a distribution over words in the vocabulary and 
documents are generated from the distribution of (latent) topics. Recent works 
showed that auto-encoders can also be seen as generative models for images and
text~\cite{Kingma:2014:AEVB,NVI:2016}. Generative models allows us to
incorporate prior information about the latent variables, and with the help
of variational Bayes (VB) 
techniques~\cite{Bishop:2006:PRML,Kingma:2014:AEVB,Rezende:2014:SBP}, one can 
infer posterior distribution over the latent variables instead of just point-
estimates. The posterior distribution captures uncertainty of the latent 
variable estimates while trying to explain (fit) the observed data and our 
prior belief. In the context of text modelling, these latent variables are seen as embeddings.


In this paper, we present Bayesian subspace multinomial model (\baysmm) as a
generative model for \bow~representation of documents. We show that our model
can learn to represent each document in the form of a Gaussian 
distribution, there by encoding the uncertainty in its covariance. 
Further, we propose a generative Gaussian classifier that exploits this 
uncertainty for topic identification (ID). The proposed VB framework can be 
extended in a straightforward way for subspace \(n\)-gram 
model~\cite{Mehdi:2013:SnGM}, that can model \(n\)-gram distribution of words 
in sentences.

Earlier, (non-Bayesian) SMM was used for learning document embeddings in an
unsupervised fashion. They were then used for training linear classifiers for  
topic ID from spoken and textual 
documents~\cite{May:2015:mivec,Kesiraju:2016:SMM}.
However, one of the limitations was that the learned document embeddings
(also termed as document i-vectors) were only point-estimates and
were prone to over-fitting, especially for shorter documents.
Our proposed model can overcome this problem by capturing the uncertainty of
the embeddings in the form of posterior distributions.


Given the significant prior research in PTMs and related algorithms for 
learning representations, it is important to draw precise relations between the 
presented model and former works. We do this from the following viewpoints: (a) 
Graphical models illustrating the dependency of random and observed variables, 
(b) assumptions of distributions over random variables and their limitations, 
and (c) approximations made during the inference and their consequences.

The contributions of this paper are as follows: (a) we present Bayesian
subspace multinomial model and analyse its relation to popular models such
as latent Dirichlet allocation (LDA)~\cite{Blei:2003:LDA}, correlated topic
model (CTM)~\cite{Blei:2005:CTM}, paragraph vector 
(PV-DBOW)~\cite{Quoc:2014:PV} and neural variational document model 
(NVDM)~\cite{NVI:2016}, (b) we adapt tricks from~\cite{Kingma:2014:AEVB} for 
faster and efficient variational inference of the proposed model, (c) we 
combine optimization techniques from~\cite{Kingma:2014:Adam,Andrew:2007:L1} and 
use them to train the proposed model, (d) we propose a generative Gaussian 
classifier that exploits uncertainty in the posterior distribution of document 
embeddings, (e) we provide experimental results on both text and speech data 
showing that the proposed document representations achieve state-of-the-art 
perplexity scores, and (f) with our proposed classification systems, we 
illustrate robustness of the model to over-fitting and at the same time obtain 
superior classification results when compared systems based on state-of-the-art 
unsupervised models.

We begin with the description of Bayesian SMM in Section~\ref{sec:baysmm},
followed by VB for the model in Section~\ref{sec:vi}. The complete VB training
procedure and algorithm is presented in Section~\ref{ssec:vb_training}. The
procedure for inferring the document embedding posterior distributions for
(unseen) documents is described in Section~\ref{ssec:extraction}.
Section~\ref{sec:glcu} presents a generative Gaussian classifier that exploits
the uncertainty encoded in document embedding posterior distributions.
Relationship between Bayesian SMM and existing popular topic models is
described in Section~\ref{sec:related_models}. Experimental details are given
in Section~\ref{sec:exp}, followed by results and analysis in
Section~\ref{sec:results}. Finally, we conclude and discuss directions for 
future research in Section~\ref{sec:concl}

\section{Bayesian subspace multinomial model}
\label{sec:baysmm}
Our generative probabilistic model assumes that the training data (\bow)
were generated as follows:

For each document, a \(K\)-dimensional latent vector \(\mb{w}\) is
generated from isotropic \Gau~prior with mean \(\mb{0}\) and precision \(\lambda\):
\begin{equation}
\label{eq_prior}
\mb{w} \sim \mathcal{N}(\mb{w} \,|\, \bs{0}\,, \mathrm{diag}((\lambda \mb{I})^{-1}))
\end{equation}
The latent vector \(\mb{w}\) is a low dimensional embedding (\(K \ll V\)) of
document-specific distribution of words, where \(V\) is the size of the 
vocabulary. More precisely, for each document, the \(V\)-dimensional vector of 
word probabilities is calculated as:
\begin{equation}
\bs{\theta} = \mathrm{softmax}(\mb{m} + \mb{T}\, \mb{w}),
\end{equation}
where \(\{\mb{m}, \mb{T}\}\) are parameters of the model. The vector \(\mb{m}\) 
known as universal background model represents log uni-gram probabilities of 
words. \(\mb{T}\) known as total variability 
matrix~\cite{Marcel:2010:SMM,Najim:2011:ivec} is a low-rank matrix defining 
subspace of document-specific distributions.

Finally, for each document, a vector of word counts \(\mb{x}\) (\bow) is sampled from \Mul~distribution:
\begin{equation}
\label{eq_sample_x}
\mb{x} \sim \mathrm{Multi}(\bs{\theta};N),
\end{equation}
where \(N\) is the number of words in the document.

The above described generative process fully defines our Bayesian model,
which we will now use to address the following problems: given training data
\(\mb{X}\), we estimate model parameters \(\{\mb{m}, \mb{T}\}\) and,
for any given document \(\widehat{\mb{x}}\), we infer posterior distribution over
corresponding document embedding \(p(\mb{w}\,|\, \widehat{\mb{x}})\).
Parameters of such posterior distribution can be then used as a low dimensional
representation of the document. Note that such distribution also encodes the
inferred uncertainty about such representation.

Using Bayes' rule, the posterior distribution of document embedding \(\mb{w}\) 
is written as\footnote{For clarity, explicit conditioning on \(\mb{T}\) and 
\(\mb{m}\) is omitted in the subsequent equations.}: 
\begin{equation}
\label{eq_ivec_post_b}
p(\mb{w} | \mb{x}) = \frac{p(\mb{x} | \mb{w})p(\mb{w})}{\int p(\mb{x} | \mb{w})p(\mb{w}) \, \dd \mb{w}}.
\end{equation}
In numerator of \eqref{eq_ivec_post_b}, \(p(\mb{w})\) represents prior 
distribution of document embeddings \eqref{eq_prior} 
and \(p(\mb{x} | \mb{w})\) represents the likelihood of observed data. 
According to our generative process, we assume that every document \(\mb{x}\) 
is a sample from \(\mathrm{Multinomial}\) distribution \eqref{eq_sample_x}, 
and the \(\log\)-likelihood is given as follows:
\begin{align}
\log p(\mb{x} | \mb{w}) &= \sum_{i=1}^{V} x_{i} \, \log \theta_{i}, \\
&= \sum_{i=1}^{V} x_{i} \log(\myfrac{\exp{m_i + \mb{t}_i \mb{w}}}{\sum_j \exp{m_j + \mb{t}_j \mb{w}}}), \\
&= \sum_{i=1}^{V} x_{i} \Bigg[ (m_i + \mb{t}_i \mb{w}) - \nonumber \\
&{} \qquad \qquad \log({\sum_{j=1}^{V} \exp{m_j + \mb{t}_j \mb{w}}}) \Bigg],  \label{eq_llh}
\end{align}
where \(\mb{t}_i\) represents a row in matrix \(\mb{T}\).
\begin{figure}[t!]
  \centering
  \captionsetup{justification=centering}
	\includegraphics[width=0.7\linewidth]{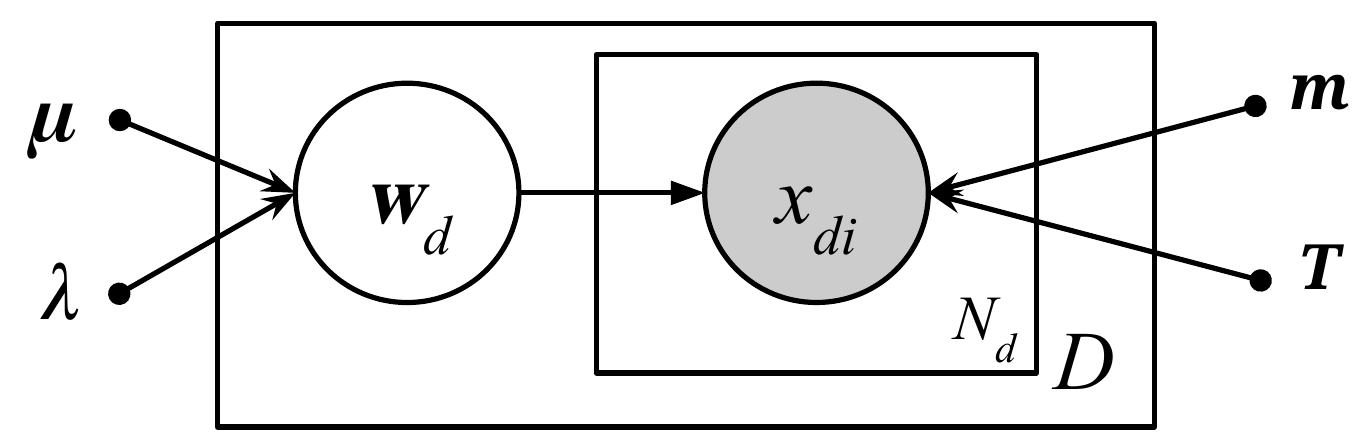}
  \caption{\label{fig:baysmm} Graphical model for Bayesian SMM}
\end{figure}
The problem arises while computing the denominator in \eqref{eq_ivec_post_b}. 
It involves solving the integral over a product of likelihood term containing 
the \(\mathrm{softmax}\) function and \Gau~distribution (prior). There exists 
no analytical form for this integral. This is a generic problem that arises 
while performing Bayesian inference for mixed-logit 
models~\cite{Blei:2005:CTM,Depraetere:2017:mixed}, multi-class 
logistic regression or any other model where likelihood function and prior are 
not conjugate to each other~\cite{Bishop:2006:PRML}. In such cases, one can 
resort to variational inference and find an approximation to the posterior 
distribution \(p(\mb{w} | \mb{x})\). This approximation to the true posterior 
is referred as variational distribution \(q(\mb{w})\), and is obtained by 
minimizing the Kullback-Leibler (KL) divergence, \(\kl(q \,||\, p)\) between 
the approximate and true posterior. We can express \(\log\) marginal (evidence) 
of the data as:
\begin{align}
\log p(\mb{x}) &=  \mathbb{E}_q[\log p(\mb{x}, \mb{w})] + \mathrm{H}[q] \, + \kl(q \,||\, p), \label{eq_log_marginal_a}   \\
{}&=  \mathcal{L}(q)  + \kl(q \,||\, p). \label{eq_log_marginal_b}
\end{align}
Here \(\mathrm{H}[q]\) represents the entropy of \(q(\mb{w})\).
Given the data \(\mb{x}\), \(\log p(\mb{x})\) is a constant with respect to \(\mb{w}\), and \(\kl(q \,||\, p)\) can be minimized by maximizing \(\mathcal{L}(q)\), which is known as \textit{Evidence Lower BOund}  (ELBO) for a document. This is the standard formulation of variational Bayes~\cite{Bishop:2006:PRML}, where the problem of finding an approximate posterior is transformed into optimization of the functional \(\mathcal{L}(q)\).
\section{Variational Bayes}
\label{sec:vi}
In this section, using the VB framework, we derive and explain the procedure 
for estimating model parameters \(\{\mb{m}, \mb{T}\}\) and inferring the 
variational distribution, \(q(\mb{w})\). Before proceeding, we note that our 
model assumes that all documents and the corresponding document embeddings 
(latent variables) are independent. This can be seen from the graphical model 
in Fig.~\ref{fig:baysmm}. Hence, we derive the inference only for one document 
embedding \(\mb{w}\), given an observed vector of word counts \(\mb{x}\).

We chose the variational distribution \(q(\mb{w})\) to be \Gau, with mean \(\bs{\nu}\) and precision \(\bs{\Gamma}\), i.e., \(q(\mb{w}) = \mathcal{N}(\mb{w} \,|\,\bs{\nu}, \, \bs{\Gamma}\I)\). The functional \(\mathcal{L}(q)\) now becomes:
%
\begin{align}
\mathcal{L}(q) &= \mathbb{E}_q[\log p(\mb{x}, \mb{w})] \,+\, \mathrm{H}[q], \\
&= -\underbrace{\kl(q \,||\, p)}_{\mathsf{A}} \,+\, 
\underbrace{\mathbb{E}_q[\log p(\mb{x} \mid \mb{w})]}_{\mathsf{B}}. 
\label{eq_elbo_b}
\end{align}
The term \(\mathsf{A}\) from \eqref{eq_elbo_b} is the negative
KL divergence between the variational distribution \(q(\mb{w})\)
and the document-independent prior from \eqref{eq_prior}. This can be 
computed analytically~\cite{cookbook} as:
\begin{multline}
\label{eq_kld_prior}
 D_{\mathrm{KL}}(q \,||\, p) = \frac{1}{2} \Big[\lambda \tr (\bs{\Gamma}^{\I}) \,+\, \log |\bs{\Gamma}| \,-\, K \log \lambda \\
+ \lambda \, \bs{\nu}^{\T} \bs{\nu} - K \Big],
\end{multline}
where \(K\) denotes the document embedding dimensionality. The term 
\(\mathsf{B}\) from \eqref{eq_elbo_b} is the expectation over
\(\log\)-likelihood of a document \eqref{eq_llh}:
\begin{multline}
\label{eq_exp_lse}
\mathbb{E}_q[\log p(\mb{x} \,|\, \mb{w})] = \sum_{i=1}^{V} x_i \Bigg[(m_i + \mb{t}_i\bs{\nu}) \\
-\underbrace{\mathbb{E}_q \Big[\log (\sum_{j=1}^V \exp{m_{j} + \mb{t}_j \mb{w}}) \Big]}_{\mathcal{F}} \Bigg].
\end{multline}
\eqref{eq_exp_lse} involves solving the expectation over \lse~operation 
(denoted by \(\mathcal{F}\)), which is intractable. It appears when dealing 
with variational inference in mixed-logit 
models~\cite{Blei:2005:CTM,Depraetere:2017:mixed}. We can approximate 
\(\mathcal{F}\) with empirical expectation using samples from \(q(\mb{w})\), 
but \(\mathcal{F}\) is a function of \(q(\mb{w})\), whose parameters we are 
seeking by optimizing \(\mathcal{L}(q)\). The corresponding gradients of 
\(\mathcal{L}(q)\) with respect to \(q(\mb{w})\) will exhibit high variance if 
we directly take samples from \(q(\mb{w})\) for the empirical expectation. To 
overcome this, we will re-parametrize the random variable 
\(\mb{w}\)~\cite{Kingma:2014:AEVB}. This is done by introducing a 
differentiable function \(g\) over another random variable \(\bs{\epsilon}\). 
If \(p(\bs{\epsilon}) = \mathcal{N}(\mb{0}, \mb{I})\), then,
\begin{equation}
\label{eq:rep}
\mb{w} = g(\bs{\epsilon}) = \bs{\nu} + \mb{L} \,\, \bs{\epsilon},
\end{equation}
where \(\mb{L}\) is the Cholesky factor of \(\bs{\Gamma}^{\I}\).
%
Using this re-parametrization of \(\mb{w}\), we obtain the following approximation:
\begin{equation}
\label{eq_exp_rep}
\mathcal{F} \approx \frac{1}{R} \sum_{r=1}^{R} \log (\sum_{j=1}^{V} \exp{m_j + \mb{t}_j \,g(\tilde{\bs{\epsilon}}_{r})}),
\end{equation}
where \(R\) denotes the total number of samples \(\tilde{\bs{\epsilon}}_r\) 
from \(p(\bs{\epsilon})\).
%
%
Combining \eqref{eq_kld_prior},\eqref{eq_exp_lse} and \eqref{eq_exp_rep}, 
we get the approximation to \(\mathcal{L}(q)\). We will introduce the document 
suffix \(d\), to make the notation explicit:
\begin{multline}
\label{eq_complete_doc_elbo}
\mathcal{L}(q_d) \approx -D_{\mathrm{KL}}(q_d \mid\mid p) + \sum_{i=1}^{V} x_{di} \,\Big[\,(m_i + \mb{t}_i\bs{\nu}_d) \\
-\frac{1}{R} \sum_{r=1}^{R} \log (\sum_{j=1}^{V} \exp{m_j + \mb{t}_j \,g(\tilde{\bs{\epsilon}}_{dr})})\Big].
\end{multline}
For the entire training data \(\mb{X}\), the complete ELBO will be simply the summation over all the documents, i.e., \(\sum_d \mathcal{L}(q_d)\).
%
%
\subsection{Training}
\label{ssec:vb_training}
The variational Bayes (VB) training procedure for Bayesian SMM is stochastic 
because of the sampling involved in the re-parametrization trick
\eqref{eq:rep}. Like the standard VB approach~\cite{Bishop:2006:PRML}, we 
optimize ELBO alternately with respect to \(q(\mb{w})\) and \(\{\mb{m}, 
\mb{T}\}\). Since we do not have closed form update equations, we perform 
gradient-based updates. Additionally, we regularize rows in matrix \(\mb{T}\) 
while optimizing. Thus, the final objective function becomes,
\begin{equation}
\label{eq_complete_reg_elbo}
\mathcal{L} = \sum_{d=1}^{D} \mathcal{L}(q_d) - \omega \sum_{i=1}^{V} ||\mb{t}_i||_{\nr{1}}\,,
\end{equation}
where we have added the term for \(\ell_1\) regularization of rows in matrix 
\(\mb{T}\), with corresponding weight \(\omega\). The same regularization was 
previously used for non Bayesian SMM in~\cite{Kesiraju:2016:SMM}. This can also 
be seen as obtaining a maximum a posteriori estimate of \(\mb{T}\) with Laplace 
priors.
\subsubsection{Parameter initialization}
\label{ssec:param_init}
The vector \(\mb{m}\) is initialized to \(\log\) uni-gram probabilities
estimated from  training data. The values in matrix \(\mb{T}\) are randomly 
initialized from \(\mathcal{N}(0, 0.001)\). The prior over latent variables 
\(p(\mb{w})\) is set to isotropic Gaussian distribution with mean \(\mb{0}\)
and \(\lambda = \{1, 10\}\).
The variational distribution \(\mb{q}(w)\) is initialized to 
\(\mathcal{N}(\mb{0}, \mathrm{diag}(0.1))\). Later in Section~\ref{ssec:conv}, 
we will show that initializing the posterior to a sharper Gaussian distribution 
helps to speed up the convergence.
%

\subsubsection{Optimization}
\label{ssec:optim}
The gradient-based updates are done by \textsc{adam} optimization scheme~\cite{Kingma:2014:Adam}; in addition to the following tricks:

We simplified the variational distribution \(q(\mb{w})\) by making its 
precision matrix \(\bs{\Gamma}\) diagonal\footnote{This is not a limitation of 
the model, but only a simplification.}. Further, while 
updating it, we used \(\log\) 
standard deviation parametrization, i.e.,
\begin{equation}
\bs{\Gamma}^{\I} = \mathrm{diag}(\exp\{2\,\bs{\varsigma}\}).
\end{equation}
The gradients of the objective \eqref{eq_complete_doc_elbo} w.r.t. the mean 
\(\bs{\nu}\) is given as follows:
\begin{align}
\label{eq_grad_nu}
\nabla \bs{\nu} &= \Big[\sum_{i=1}^{V} \mb{t}_i^{\T}(x_i - \frac{1}{R}\sum_{r=1}^{R}\theta_{ir} \sum_{k=1}^{V}x_k )  \Big] - \lambda \bs{\nu} \\
\intertext{where,}
\theta_{ir} &= \myfrac{\exp\{m_i + \mb{t}_jg(\bs{\epsilon}_r)\}}{\sum_j \exp\{m_j + \mb{t}_jg(\bs{\epsilon}_r)\}} \label{eq_theta_ir}
\end{align}
The gradient w.r.t \(\log\) standard deviation \(\bs{\varsigma}\) is given as:
\begin{multline}
\label{eq_grad_lstd}
\nabla \bs{\varsigma} =  \mb{1} - \lambda \exp\{2 \bs{\varsigma}\} \, \\
- \sum_{k=1}^{V} x_k\frac{1}{R}\sum_{r=1}^{R}\sum_{i=1}^{V}\theta_{ir} \mb{t}_i^{\T} \odot \exp\{\bs{\varsigma}\} \odot \bs{\epsilon}_r,
\end{multline}
where \(\mb{1}\) represents a column vector of ones, \(\odot\) denotes 
element-wise product, and \(\exp\) is element-wise exponential operation.

The \(\ell_1\) regularization term makes the objective function 
\eqref{eq_complete_reg_elbo} discontinuous (non-differentiable) at points 
where it crosses the orthant. Hence, we used sub-gradients and employed 
orthant-wise learning~\cite{Andrew:2007:L1}. The gradient of the objective 
\eqref{eq_complete_reg_elbo} w.r.t. a row \(\mb{t}_i\) in matrix \(\mb{T}\) is 
computed as follows:
\begin{multline}
\label{eq_grad_t}
\nabla \mb{t}_i = -\omega \, \mathrm{sign}(\mb{t}_i) + \sum_{d=1}^{D} \Bigg[x_{di} \bs{\nu}_d^{\T} - \\
\Big[\Big(\sum_{k=1}^{V} x_{ki} \Big) \frac{1}{R}\sum_{r=1}^{R}\theta_{dir} (\bs{\nu}_d^{\T} + \bs{\epsilon}_{dr}^{\T} \odot \exp\{\bs{\varsigma}^{\T}\}) \Big]  \Bigg].
\end{multline}
Here, \(\mathrm{sign}\) and \(\exp\) operate element-wise. The sub-gradient \(\tilde{\nabla} \mb{t}_i\) is defined as:
\begin{equation}
\label{eq_subg}
\tilde{\nabla} {t}_{ik} \triangleq \left\{
\arraycolsep=3pt\def\arraystretch{1.1}
\begin{array}{lcccl}
\nabla {t}_{ik} + \omega, & \quad t_{ik}   & = 0, & \nabla {t}_{ik} & < -\omega \\
\nabla {t}_{ik} - \omega, & \quad t_{ik}   & = 0, & \nabla {t}_{ik} & > \omega \\
0,                        & \quad t_{ik}   & = 0, & |\nabla {t}_{ik}| & \leq \omega \\
\nabla {t}_{ik},          & \quad |t_{ik}| & > 0  & .
\end{array} \right.
\end{equation}
Finally, the rows in matrix \(\mb{T}\) are updated according to,
\begin{align}
\label{eq_t_update}
\mb{t}_i &\leftarrow \mathcal{P_O}(\mb{t}_i + \mb{d}_i)\, \\
\intertext{where, \(\mb{d}_i\) is the step in ascent direction,}
\mb{d}_i &= \eta \, \mathrm{diag}(\sqrt{\mb{s}_i} + \epsilon)^{\I} \, 
\mb{f}_{i} \label{eq_t_step}\,. 
\end{align}
Here, \(\eta\) is the learning rate, \(\mb{f}_i\) and \(\mb{s}_i\) represents 
bias corrected first and second moments (as required by \textsc{adam}) of 
sub-gradient \(\tilde{\nabla}\mb{t}_{i}\) respectively.
\(\mathcal{P}_{O}\) represents orthant projection, which ensures that the 
update step does not cross the point of non-differentiability. It is defined as,
\begin{equation}
\mathcal{P_O}(\mb{t}_i + \mb{d}_i) \triangleq \left\{
\begin{array}{ll} 0  & \textrm{if} \quad t_{ik}(t_{ik} + d_{ik}) < 0, \\
t_{ik} + d_{ik} & \textrm{otherwise} \,.
\end{array} \right. \label{eq_orth}
\end{equation}
The orthant projection introduces explicit zeros in the estimated \(\mb{T}\) 
matrix and, results in sparse solution. Unlike in~\cite{Kesiraju:2016:SMM}, we 
do not require to apply the sign projection, because both the gradient
\(\tilde{\nabla}\mb{t}_{i}\) and step \(\mb{d}\) point to the same orthant (due 
to properties of \textsc{adam}).
%
The stochastic VB training is outlined in Algorithm~\ref{alg_vb}.
%
%
\begin{algorithm}[!t]
	\DontPrintSemicolon
	initialize the model and the variational parameters\;
	\Repeat{convergence or max\_iterations}{
		\For{\(d=1 \ldots D\,\)}{
			sample \(\tilde{\bs{\epsilon}}_{dr} \sim \mathcal{N}(\mb{0}, \mb{I}) \qquad r=1\ldots R\)\;
			compute \(\mathcal{L}(q_d)\) using \eqref{eq_complete_doc_elbo}\;
			compute gradient \(\nabla \bs{\nu}_d\) using \eqref{eq_grad_nu}\;
			compute gradient \(\nabla \bs{\varsigma}_d\) using \eqref{eq_grad_lstd}\;
			update \(\bs{\nu}_d\) and \(\bs{\varsigma}_d\) using \textsc{adam} \;
		}
		compute \(\mathcal{L}\) using \eqref{eq_complete_reg_elbo} \;
		compute sub-gradients \(\tilde{\nabla}\mb{t}_i\) using 
		\eqref{eq_grad_t} and \eqref{eq_subg}\;
		update rows in \(\mb{T}\) using \eqref{eq_t_update}\;
	}
	\caption{Stochastic VB training} \label{alg_vb}
\end{algorithm}
%
%
\subsection{Inferring embeddings for new documents}
\label{ssec:extraction}
After obtaining the model parameters from VB training, we can infer (extract) 
the posterior distribution of document embedding \(q(\mb{w})\) for any given 
document \(\mb{x}\). This is done by iteratively updating the parameters of 
\(q(\mb{w})\) that maximize \(\mathcal{L}(q)\) from 
\eqref{eq_complete_doc_elbo}. These updates are performed by following the 
same \textsc{adam} optimization scheme as in training.

Note that the embeddings are extracted by maximizing the ELBO, that does not 
involve any supervision (topic labels). These embeddings which are in the form 
of posterior distributions will be used as input features for training topic ID 
classifiers. Alternatively, one can use only the mean of the posterior 
distributions as point estimates of document embeddings.

%
\section{Gaussian classifier with uncertainty}
\label{sec:glcu}
In this section, we will present a generative Gaussian classifier that exploits 
the uncertainty in posterior distributions of document embedding. Moreover, it 
also exploits the same uncertainty while computing the posterior probability of 
class labels. The proposed classifier is called Gaussian linear classifier with 
uncertainty (GLCU) and is inspired by~\cite{Kenny:2013:PLDA,Sandro:2015:IS}. 
It can be seen as an extension to the simple Gaussian linear classifier 
(GLC)~\cite{Bishop:2006:PRML}. 

Let \(\ell = 1\ldots L\) denote class labels, \(d=1\ldots D\) represent document indices, and \(\mb{h}_d\) represent the class label of document \(d\) in one-hot encoding.

GLC assumes that every class is Gaussian distributed with a specific mean \(\bs{\mu}_{\ell}\), and a shared precision matrix \(\mb{D}\). Let \(\mb{M}\) denote a matrix of class means, with \(\bs{\mu}_{\ell}\) representing a column. GLC is described by the following model:
\begin{equation}
\label{eq_glc_model}
\mb{w}_d = \bs{\mu}_{d} + \bs{\varepsilon}_d, \\
\end{equation}
where \(\bs{\mu}_{d}=\mb{M} \mb{h}_d\), \(p(\bs{\varepsilon})= \mathcal{N}(\bs{\varepsilon}\,|\, \mb{0}, \mb{D}^{\I})\) and \(\mb{w}_d\) represent embedding for document \(d\). GLC can be trained by estimating the parameters \(\varTheta=\{\mb{M}, \mb{D}\}\) that maximize the class conditional likelihood of all training examples:
\begin{equation}
\label{eq_cc_llh_glc}
\prod_{d=1}^{D} p(\mb{w}_d \,|\, \mb{h}_d, \varTheta) = \prod_{d=1}^{D} \mathcal{N}(\mb{w}_d \,|\, \bs{\mu}_{d}, \mb{D}^{\I}).
\end{equation}
In our case, however, the training examples come in the form of posterior distributions, \(q(\mb{w}_d) = \mathcal{N}(\mb{w}_d\,|\,\bs{\nu}_d, \bs{\Gamma}_d^{\I})\) as extracted using our Bayesian SMM. In such case, the proper ML training procedure should maximize the expected class-conditional likelihood, with the expectation over \(\bs{w}_d\) calculated for each training example with respect to its posterior distribution \(q(\mb{w}_d)\) i.e., \(\mathbb{E}_q[\mathcal{N}(\mb{w}_d \,|\, \bs{\mu}_d, \mb{D}^{\I})]\).

However, it is more convenient to introduce an equivalent model, where the 
observations are the means \(\bs{\nu}_d\) of the posteriors \(q(\mb{w}_d)\) and 
the uncertainty encoded in \(\bs{\Gamma}_d^{\I}\) is introduced into the model 
through the latent variable \(\mb{y}_d\) as,
\begin{equation}
\label{eq_clf_model}
\bs{\nu}_d = \bs{\mu}_{d} + \mb{y}_d + \bs{\varepsilon}_d,
\end{equation}
where, \(p(\mb{y}_d)=\mathcal{N}(\mb{y}_d \,|\, \mb{0}, \bs{\Gamma}_{d}^{\I})\).
The resulting model is called GLCU. Since the random variables \(\mb{y}_d\) and 
\(\bs{\epsilon}_d\) are Gaussian-distributed, the resulting
class conditional likelihood is obtained using convolution of two 
Gaussians~\cite{Bishop:2006:PRML}, i.e,
\begin{equation}
\label{eq_cc_llh_a}
p(\bs{\nu}_d \,|\, \mb{h}_d, \varTheta) =
\mathcal{N}(\bs{\nu}_d \,|\, \bs{\mu}_d,\, \bs{\Gamma}_{d}^{\I}+\mb{D}^{\I}).
\end{equation}
GLCU can be trained by estimating its parameters \(\varTheta\), that maximize 
the class conditional likelihood of training data \eqref{eq_cc_llh_a}. 
This can be done efficiently by using the following \textsc{em} algorithm. 

\subsection{EM algorithm}
In the \textsc{e}-step, we calculate the posterior distribution of latent 
variables: 
\begin{align}
\label{eq_e_step_clf}
p(\mb{y}_d \,|\, \bs{\nu}_d, \mb{h}_d, \varTheta) &\propto p(\bs{\nu}_d \,|\, 
\mb{y}_d, \mb{h}_d\varTheta) \,\, p(\mb{y}_d) \nonumber \\
&\propto \, \mathcal{N}(\mb{y}_d \,|\,\mb{u}_d, \mb{V}^{\I}_d),
\intertext{where,}
\mb{V}_d &= \mb{D} + \bs{\Gamma}_d, \\
\mb{u}_d &= [\mb{I} + \mb{D}^{\I} \bs{\Gamma}_d]^{\I} (\bs{\nu}_d-\bs{\mu}_{d}).
\end{align}

In the \textsc{m}-step, we maximize the auxiliary function \(\mathcal{Q}\) with respect to model parameters \(\varTheta\). It is the expectation of \(\log\) joint-probability with respect to \(p(\mb{y}_d \,|\, \bs{\nu}_d)\), i.e.,
\begin{align}
\mathcal{Q} &= \mathbb{E}_{p}[\sum_{d=1}^D \log p(\bs{\nu}_d, \mb{y}_d \mid \varTheta)] \\
&= \frac{-D}{2}\log |\mb{D}| - \frac{1}{2} \Bigg[\sum_{d=1}^{D} 
\Big(\mathrm{tr}(\mb{D}\mb{V}_d^{\I})
\nonumber \\
&\,+ \big(\mb{u}_d-(\bs{\nu}_d-\bs{\mu}_{d})\big)^{\T} \mb{D} 
\big(\mb{u}_d-(\bs{\nu}_d-\bs{\mu}_{d})\big) \Big) \Bigg] + \mathrm{const}.
\end{align}
Maximizing the auxiliary function \(\mathcal{Q}\) w.r.t. \(\varTheta\), we have:
\begin{align}
\bs{\mu}_{\ell} &:= \frac{1}{|\mathcal{I}_{\ell}|} \sum_{d \in \mathcal{I}_{\ell}} (\bs{\nu}_d -
\mb{u}_d) \label{eq_mu_l_update} \quad \forall \, \ell=1\ldots L \\
\mb{D}^{\I} &:= \frac{1}{D} \Big[\sum_{d=1}^{D} (\mb{a}_{d} \, \mb{a}^{\T}_{d}) 
+ \mb{V}_d^{\I} \Big],
\end{align}
where, \(\mb{a}_d = \mb{u}_d - (\bs{\nu}_d - \bs{\mu}_d)\), and, 
\(\mathcal{I}_{\ell}\) is the set of documents from class \(\ell\).
To train the GLCU model, we alternate between \textsc{e}-step and \textsc{m}-step until convergence.


%
\subsection{Classification}
\label{ssec:gc_pred}
Given a test document embedding posterior distribution \(q(\mb{w}) = 
\mathcal{N}(\mb{w}\,|\,\bs{\nu}, \bs{\Gamma}^{\I})\), we compute the class 
conditional likelihood according to \eqref{eq_cc_llh_a}, and the posterior 
probability of a class \(\mathcal{C}_{k}\) is obtained by applying the Bayes' 
rule:
\begin{align}
\label{eq_cc_llh}
p(\mathcal{C}_{k} \,|\, \bs{\nu}, \bs{\Gamma}, \varTheta) = \myfrac{p(\bs{\nu} 
\mid \bs{\mu}_{k}, \mb{D}, \bs{\Gamma}) \, p(\mathcal{C}_{k})}{\sum_{\ell} 
p(\bs{\nu} \mid \bs{\mu}_{\ell}, \mb{D}, \bs{\Gamma}) \, p(\mathcal{C}_{\ell})} 
.
\end{align}

\section{Related models}
\label{sec:related_models}
In this section, we review and relate some of the popular PTMs and neural network based document models. We begin with a brief review of LDA~\cite{Blei:2003:LDA}, a probabilistic generative model for \bow~representation of documents.

\subsection{Latent Dirichlet allocation}
Let \(\bs{\phi}_{1:K}\) represent \(K\) topics. LDA assumes that every topic 
\(\bs{\phi}_k\) is a distribution over a fixed vocabulary of size \(V\). Every
document \(d\) is generated by a two step process:
First, a document-specific vector (embedding) representing a distribution over 
\(K\) topics is sampled, i.e., \(\bs{\theta}_d \sim 
\mathrm{Dir}(\bs{\alpha})\). 
Then, for each word in the document \(d\), a topic indicator variable
\(z_{i}\) is sampled: \(z_i \sim \mathrm{Multi}(\bs{\theta}_d; 1)\) and the 
word \(x_{i}\) is in turn sampled from the topic-specific distribution: \(x_{i} 
\sim \mathrm{Multi}(\bs{\phi}_{z_{i}}; 1)\).

\begin{figure}[t!]
	\centering
	\includegraphics[width=0.7\linewidth]{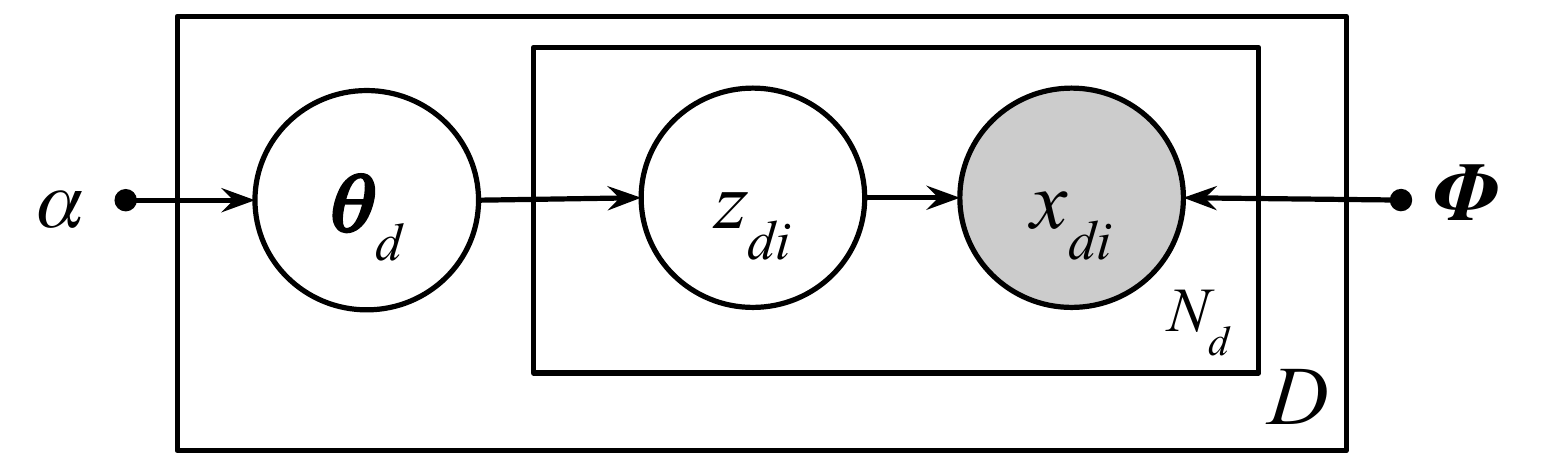}
	\caption{Graphical model for LDA} \label{fig:lda}
\end{figure}
The topic (\(\bs{\phi}\)) and document (\(\bs{\theta}\)) vectors live in \((V-1)\) and \((K-1)\) simplexes respectively. For every word \(x_{i}\) in document \(d\), there is a discrete latent variable \(z_{i}\) that tells which topic was responsible for generating the word. This can be seen from the respective graphical model in Fig.~\ref{fig:lda}.
%

During inference, the generative process is inverted to obtain posterior 
distribution over latent variables, \(p(\bs{\theta},\mb{z}\,|\, 
\mb{x},\bs{\alpha}, \bs{\phi}_{1:K})\), given the observed data and prior 
belief.
Since the true posterior is intractable, Blei~\cite{Blei:2003:LDA} resorted to 
variational inference which finds an approximation to the true posterior as a 
variational distribution \(q(\bs{\theta}, \mb{z})\). Further, mean-field 
approximation was made to make the inference tractable, i.e., \(q(\bs{\theta}, 
\mb{z}) = q(\bs{\theta})\prod_{i}q(z_i)\).

In the original model proposed by Blei~\cite{Blei:2003:LDA}, the parameters \(\bs{\phi}\) were obtained using maximum likelihood approach. The choice of \Dir~distribution for \(q(\bs{\theta})\)  simplifies the inference process because of the \Dir-\Mul~conjugacy. However, the assumption of \Dir~distribution causes limitations to the model, and \(q(\bs{\theta})\) cannot capture correlations between topics in each document.
This was the motivation for Blei~\cite{Blei:2005:CTM} to model documents with \Gau~distributions, and the resulting model is called correlated topic model (CTM).

\subsection{Correlated topic model}
The generative process for a document in CTM~\cite{Blei:2005:CTM} is same as in 
LDA, except for document vectors are now drawn from \Gau, i.e.,
\begin{align}
p(\bs{\eta})  &= \mathcal{N}(\bs{\eta} \mid \bs{\mu}, \mathrm{diag}(\lambda)^{\I}), \\
\bs{\theta} &= \mathrm{softmax}(\bs{\eta}). \label{eq_ctm_softmax}
\end{align}
In this formulation, the document embeddings \(\bs{\eta}\) are no longer in the 
\((K-1)\) simplex, rather they are dependent through the logistic normal. This 
is the same as in our proposed Bayesian SMM \eqref{eq_prior}. The advantage is 
that the document vectors can model the correlations in topics. The topic 
distributions over vocabulary \(\bs{\phi}\), however, still remained \Dis.
In Bayesian SMM, the topic-word distributions (\(\mb{T}\)) are not \Dis~, hence 
it can model the correlations between words and (latent) 
topics~\cite{Blei:2005:CTM}.
\begin{figure}[t!]
	\centering
	\includegraphics[width=0.7\linewidth]{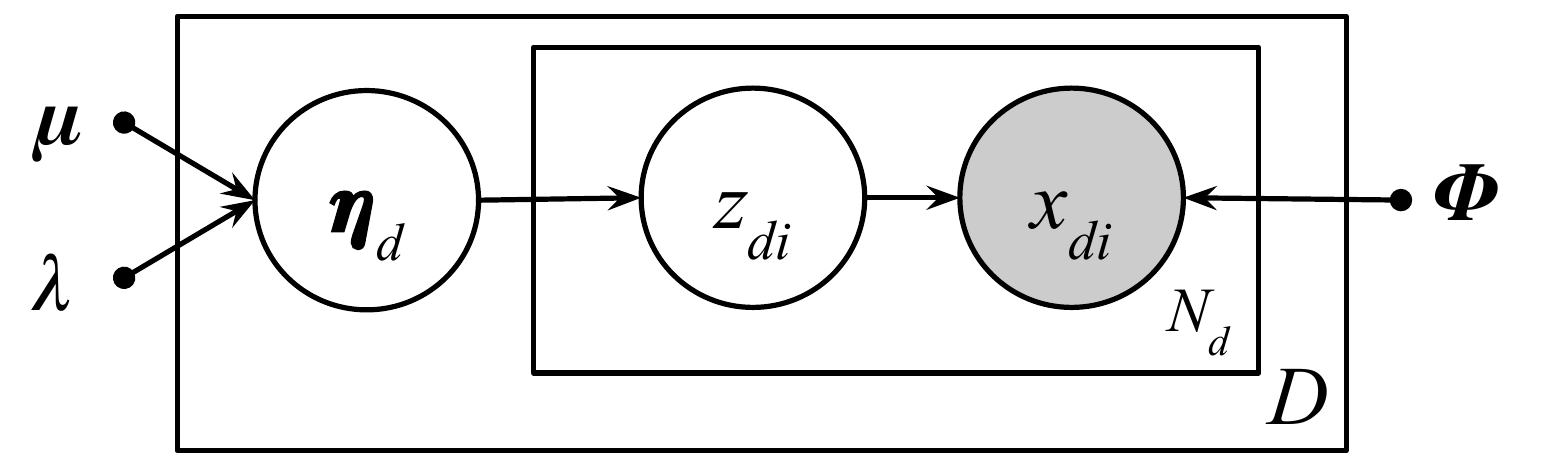}
	\caption{Graphical model for CTM} \label{fig:ctm}
\end{figure}
%

The variational inference in CTM is similar to that of LDA including the 
mean-field approximation, because of the discrete latent variable \(\mb{z}\) 
(Fig.~\ref{fig:ctm}). An additional problem is dealing with the non-conjugacy. 
More specifically, it is the intractability while solving the expectation over 
\lse~function (see \(\mathcal{F}\) from \eqref{eq_exp_lse}). 
Blei~\cite{Blei:2005:CTM} used Jensen's inequality to form an upper bound on 
\(\mathcal{F}\), and this in-turn acted as lower bound on ELBO. In our proposed 
Bayesian SMM, we also encountered the same problem, and we approximated 
\(\mathcal{F}\) using  the re-parametrization trick (Section \ref{sec:vi}). 
There exist similar approximation techniques based on Quasi Monte Carlo 
sampling~\cite{Depraetere:2017:mixed}.


Unlike in LDA or CTM, Bayesian SMM does not require to make mean-field approximation, because the topic-word mixture is not \Dis~thus eliminating the need for discrete latent variable \(\mb{z}\).
\subsection{Subspace multinomial model}
\label{ssec_smm}
SMM is a log-linear model; originally proposed for modelling discrete prosodic 
features for the task of speaker verification~\cite{Marcel:2010:SMM}. Later, it 
was used for phonotatic language recognition~\cite{Mehdi:2011:SMM} and 
eventually for topic identification and document 
clustering~\cite{May:2015:mivec,Kesiraju:2016:SMM}. Similar model was proposed 
by Maas~\cite{Maas:2011:Sent} for unsupervised learning of word 
representations. One of the major differences among these works is the type of 
regularization used for matrix \(\mb{T}\).

Another major difference is in obtaining embeddings \(\mb{w}_d\) for a given test document. Maas~\cite{Maas:2011:Sent} obtained them by projecting the vector of word counts \(\mb{x}_d\) onto the matrix \(\mb{T}\), i.e., \(\mb{w}_d = \mb{T\,x}_d\), whereas~\cite{May:2015:mivec,Kesiraju:2016:SMM} extracted the embeddings by maximizing regularized \(\log\)-likelihood function.
The embeddings extracted using SMM are prone to over-fitting. Our Bayesian SMM 
overcomes this problem by capturing the uncertainty of document embeddings in 
the posterior distribution. Our experimental analysis in 
section~\ref{ssec:ease_of_training} illustrates the robustness of Bayesian SMM.

\subsection{Paragraph vector}
\label{ssec:pvec}
Paragraph vector bag-of-words (PV-DBOW)~\cite{Quoc:2014:PV} is also a 
log-linear model, which is trained stochastically to maximize the likelihood of 
a set of words from a given document. SMM can be seen as a special case of 
PV-DBOW, since it maximizes the likelihood of all the words in a document.

\subsection{Neural network based models}
\label{ssec:nvdm}
Neural variational document model (NVDM) is an adaptation of variational auto-encoders for document modelling~\cite{NVI:2016}. The encoder models the posterior distribution of latent variables given the input, i.e., \(p_{\theta}(\mb{z} \,|\, \mb{x})\), and the decoder models distribution of input data given the latent variable, i.e., \(p_{\theta}(\mb{x} \,|\, \mb{z})\).  In NVDM, the authors used \bow~as input, while their encoder and decoders are two-layer feed-forward neural networks. The decoder part of NVDM is similar to Bayesian SMM, as both the models maximize expected \(\log\)-likelihood of data, assuming \Mul~distribution. In simple terms, Bayesian SMM is a decoder with a single feed forward layer. For a given test document, in NVDM, the approximate posterior distribution of latent variables is obtained directly by forward propagating through the encoder; whereas in Bayesian SMM, it is obtained by iteratively optimizing ELBO.
The experiments in Section~\ref{sec:results} show that the posterior distributions obtained from Bayesian SMM represent the data better as compared to the ones obtained directly from the encoder of NVDM. 

\subsection{Sparsity in topic models}
\label{sec_sparsity_tm}
Sparsity is often one of the desired properties in topic 
models~\cite{SAGE:2011,Biksha:2007:Sparse}. Sparse coding inspired topic model 
was proposed by~\cite{Zhu:2011:STC}, where the authors have obtained sparse 
representations for both documents and words. \(\ell_1\) regularization over 
\(\mb{T}\) for SMM (\(\ell_1\) SMM) was observed to yield better results when 
compared to LDA, STC and \(\ell_2\) regularized SMM (\(\ell_2\) 
SMM)~\cite{Kesiraju:2016:SMM}. Relation between SMM and sparse additive 
generative model (SAGE)~\cite{SAGE:2011} was explained in~\cite{May:2015:mivec}.
In~\cite{Mekala:2017:SCDV}, the authors proposed an algorithm to obtain sparse 
document embeddings (called sparse composite document vector (SCDV)) from 
pre-trained word embeddings. In our proposed Bayesian SMM, we introduce 
sparsity into the model parameters \(\mb{T}\) by applying \(\ell_1\) 
regularization and using orthant-wise learning.

\section{Experiments}
\label{sec:exp}
\subsection{Datasets}
\label{ssec:data}
We have conducted experiments on both speech and text corpora. The speech 
data used is \textit{Fisher} phase 1 
corpus\footnote{\texttt{\url{https://catalog.ldc.upenn.edu/LDC2004S13}}}, 
which is a collection of 5850 conversational telephone speech recordings with
a closed set of 40 topics. Each conversation is approximately 10 minutes long
with two sides of the call and is supposedly about one topic.
We considered each side of the call (recording) as an independent document,
which resulted in a total of 11700 documents. Table~\ref{tab:data} presents
the details of data splits; they are the same as used in earlier
research~\cite{Hazen:2007:ASRU,Hazen:2011:MCE_topic_ID,May:2015:mivec}. Our
preprocessing involved removing punctuation and special characters, but we
did not remove any stop words. Using Kaldi open-source
toolkit~\cite{Kaldi:2011:ASRU}, we trained a sequence discriminative DNN-HMM
automatic speech recognizer (ASR) system~\cite{Karel:2013} to obtain
automatic transcriptions. The ASR system resulted in 18\% word-error-rate on
a held-out test set. We report experimental results on both manual and
automatic transcriptions. The vocabulary size while using manual
transcriptions was 24854, for automatic, it was 18292, and the average
document length is 830, and 856 words respectively.
\begin{table}[t!]
  \begin{center}
    \caption{\label{tab:data}Data splits from \textit{Fisher} phase 1 corpus, 
    where each document represents one side of the conversation.}
    \begin{tabular}{lcc} \toprule
      \bf{Set} & \bf{\# docs.} & \bf{Duration (hrs.)}  \\ \midrule
      ASR training      & 6208 & 553 \\
      Topic ID training & 2748 & 244 \\
      Topic ID test     & 2744 & 226 \\ \bottomrule
    \end{tabular}
  \end{center}
\end{table}

The text corpus used is 
\textit{20Newsgroups}\footnote{\texttt{\url{http://qwone.com/~jason/20Newsgroups/}}},
which contains 11314 training and 7532 test documents 
over 20 topics. Our preprocessing involved removing punctuation and 
words that do not occur in at least two documents, which resulted in a 
vocabulary of 56433 words. The average document length is 290 words.

\subsection{Hyper-parameters of Bayesian SMM}
\label{ssec:hyper}
In our topic ID experiments, we observed that the embedding dimension (\(K\))
and regularization weight (\(\omega\)) for rows in matrix \(\mb{T}\) are the 
two important hyper-parameters. The embedding dimension was chosen from 
\(K=\{100, \ldots, 800\}\), and regularization weight from \(\omega=\{0.0001, 
\ldots, 10.0\}\). 

\subsection{Proposed topic ID systems}
\label{ssec:proposed_sys}
Our Bayesian SMM is an unsupervised model trained iteratively by 
optimizing the ELBO; it does not necessarily correlate
with the performance of topic ID. It is valid for SMM, NVDM or any other 
generative model trained without supervision. A typical way to overcome this 
problem is to have an early stopping mechanism (ESM), which requires to 
evaluate the topic ID accuracy on a held-out (or cross-validation) set at 
regular intervals during the training. It can then be used to stop the 
training earlier if needed.

Using the above described scheme, we trained three different classifiers:
(i) Gaussian linear classifier (GLC), (ii) multi-class logistic 
regression (LR), and, (iii) Gaussian linear classifier with uncertainty 
(GLCU). Note that GLC and LR cannot exploit the uncertainty in the document 
embeddings; and are trained using only the mean parameter \(\bs{\nu}\) of the 
posterior distributions; whereas GLCU is trained using the full posterior 
distribution \(q(\mb{w})\), i.e., along with the uncertainties of document 
embeddings as described in Section~\ref{sec:glcu}. GLC and GLCU does not have 
any hyper-parameters to tune, while the \(\ell_2\) regularization weight of LR 
was tuned using cross-validation experiments.

%
%
\subsection{Baseline topic ID systems}
\label{ssec:baselines}
\subsubsection{NVDM}
\label{ssec:nvdm_base}
Since NVDM and our proposed Bayesian
SMM share similarities, we chose to extract the embeddings from NVDM and use
them for training linear classifiers. Given a trained NVDM model, embeddings 
for any test document can be extracted just by forward propagating through 
the encoder. Although this is computationally cheaper, one needs to decide 
when to stop training, as a fully converged NVDM may not yield optimal 
embeddings for discriminative tasks such as topic ID. Hence, we used the same
early stopping mechanism as described in earlier section. We used the same 
three classifier pipelines (LR, GLC, GLCU) as we used for Bayesian SMM. Our 
architecture and training scheme are similar to ones proposed 
in~\cite{NVI:2016}, i.e., two 
feed forward layers with either \(500\) or \(1000\) hidden units and 
\(\{\mathrm{sigmoid}, \mathrm{ReLU}, \mathrm{tanh}\}\) activation functions. 
The latent dimension was chosen from \(K=\{100, \ldots, 800\}\). The 
hyper-parameters were tuned based on cross-validation experiments.

\subsubsection{SMM}
Our second baseline system is non-Bayesian SMM with \(\ell_1\) regularization
over the rows in \(\mb{T}\) matrix, i.e., \(\ell_1\) SMM. It was trained with
hyper-parameters such as embedding dimension \(K=\{100, \ldots, 
800\}\), and regularization weight \(\omega=\{0.0001, \ldots, 10.0\}\). The 
embeddings obtained from SMM were then used to train GLC and LR classifiers. 
Note that we cannot use GLCU here, because SMM yields only point-estimates of
embeddings. We used the same early stopping mechanism to train the classifiers. 
The experimental analysis in Section~\ref{ssec:ease_of_training} shows that 
Bayesian SMM is more robust to over-fitting when compared to SMM and NVDM, 
and does not require an early stopping mechanism.

\subsubsection{ULMFiT}
The third baseline system is the universal language model fine-tuned for 
classification (ULMFiT)~\cite{Ruder:2018:Universal}.
The pre-trained\footnote{\texttt{\url{https://github.com/fastai/fastai}}} 
model consists of 3 BiLSTM layers. Fine-tuning the model involves two steps: 
(a) fine-tuning LM on the target dataset and (b) training classifier (MLP 
layer) on the target dataset. We trained several models with various drop-out 
rates. More specifically, the LM was fine-tuned for 15 
epochs\footnote{Fine-tuning LM for higher number of epochs degraded the 
classification performance.}, with drop-out rates from: \(\{0.2, \ldots , 0.6\} 
\). The classifier was fine-tuned for 50 epochs with drop-out rates from: 
\(\{0.2,\ldots , 0.6\} \). A held-out development set was used to tune the 
hyper-parameters (drop-out rates, and fine-tuning epochs).

\subsubsection{TF-IDF}
The fourth baseline system is a standard term frequency-inverse document 
frequency (TF-IDF) based document representation, followed by multi-class 
logistic regression (LR). Although TF-IDF is not a topic model, the 
classification performance of TF-IDF based systems are often close to 
state-of-the-art systems~\cite{May:2015:mivec}. The hyper-parameter
(\(\ell_2\) regularization weight) of LR was selected based on 5-fold 
cross-validation experiments on training set.

\section{Results and Discussion}
\label{sec:results}

\begin{figure}[t!]
  \centering
  \includegraphics[width=0.9\linewidth]{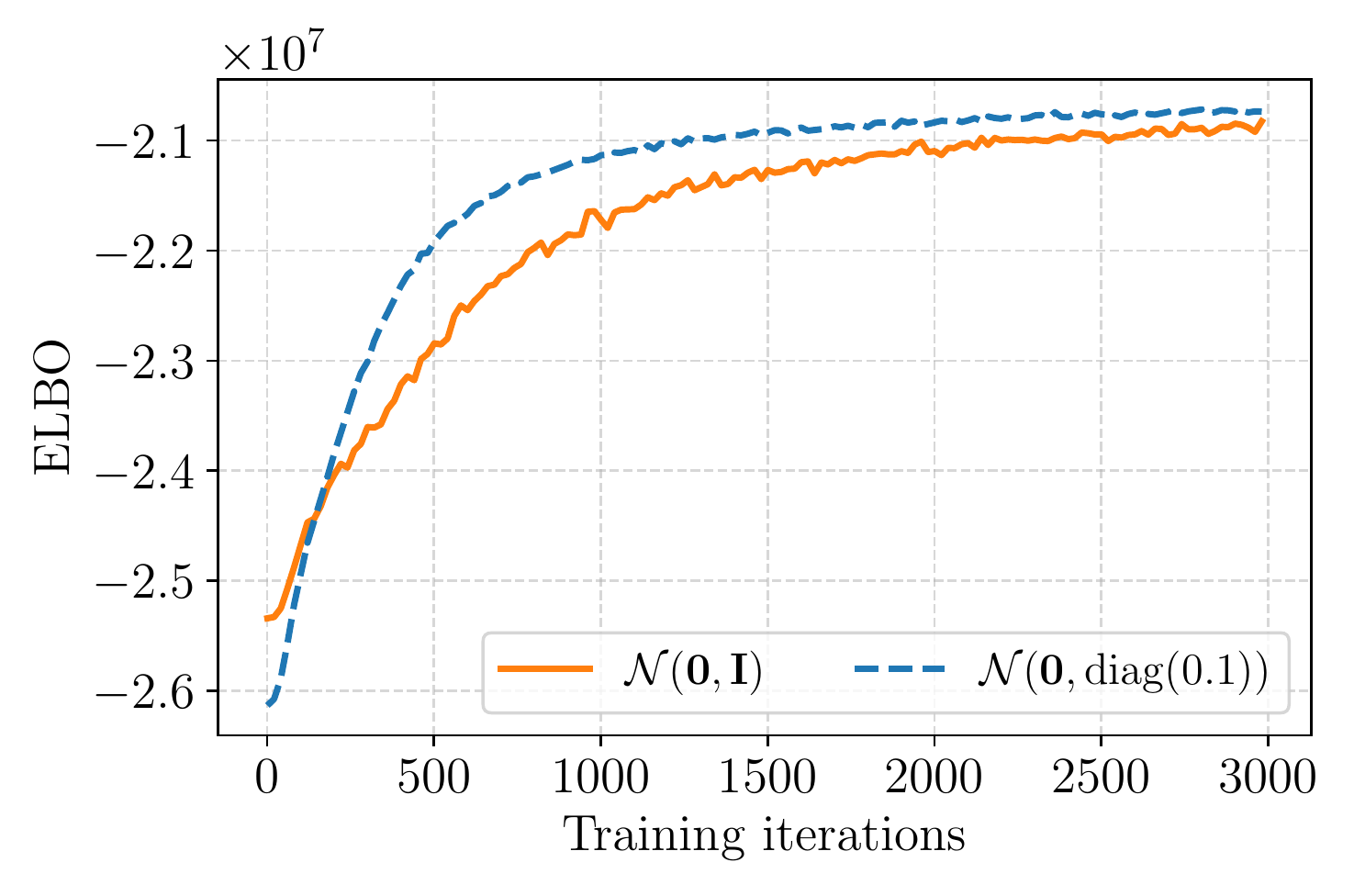}
  \caption{\label{fig:conv} Convergence of Bayesian SMM for various 
    initializations of variational distribution. The model was trained on 
    \textit{20Newsgroups} corpus with \(K = 100\), and \(\omega=1\).}
\end{figure}
\subsection{Convergence rate of Bayesian SMM}
\label{ssec:conv}
We observed that the posterior distributions extracted using Bayesian SMM are
always much sharper than standard Normal distribution. Hence we initialized 
the variational distribution to \(\mathcal{N}(\mb{0},\, \mathrm{diag}(0.1))\) 
to speed up the convergence. Fig.~\ref{fig:conv} shows objective (ELBO) plotted 
for two different initializations of variational distribution. Here, the model 
was trained on \textit{20Newsgroups} corpus, with the embedding dimension 
\(K=100\), regularization weight \(\omega=1.0\) and prior 
set to standard \Nor. We can observe that the model initialized to 
\(\mathcal{N}(\mb{0}, \, \mathrm{diag}(0.1))\)
converges faster as compared to the one initialized to standard \Nor. In all 
the further experiments, we initialized\footnote{One can introduce hyper-priors 
and learn the parameters of prior distribution.} both the prior and variational 
distributions to \(\mathcal{N}(\mb{0}, \,\mathrm{diag}(0.1))\).


\subsection{Perplexity}
Perplexity is an intrinsic measure for topic 
models~\cite{DBM:2013:Hinton,NVI:2016}. It is computed as an average of 
every test document according to:
\begin{align}
\textrm{PPL\textsubscript{DOC}} &= \exp\Big\{\myfrac{-1}{D} \sum_{d=1}^{D} 
\frac{\log p(\mb{x}_d)}{N_d}\Big\}, \\
\intertext{or for an entire test corpus according to:}
\textrm{PPL\textsubscript{CORPUS}} &= \exp\Big\{-\frac{\sum_{d=1}^{D} \log 
p(\mb{x}_d)}{\sum_{d=1}^{D} N_d}\Big\},
\end{align}
where \(N_d\) is the number of word tokens in document \(d\).

\begin{table}[t!]
	\begin{center}
		\caption{\label{tab:ppl}Comparison of perplexity (PPL) results on 
			\textit{20Newsgroups}. The values in the brackets indicate results 
			with 
			a limited vocabulary of 2000 words.}
		\begin{tabular}{lrrr} \toprule
			\textbf{Model}  & \(K\) & PPL\textsubscript{CORPUS} & 
			PPL\textsubscript{DOC} \\ \midrule
			NVDM         & 50   &  1287 (769) & 1421 (820)  \\
			NVDM         & 200  &  1387 (852) & 1519 (870)  \\
			Bayesian SMM & 50   & \textbf{1043 (629)} & \textbf{1064 (639)}  \\
			Bayesian SMM & 200  & \textbf{882 (519)} & \textbf{851 (515)}  \\ 
			\midrule
			ML estimate  &  -   & 153 (90)  & 93 (42) \\ \bottomrule
		\end{tabular}
	\end{center}
\end{table}

In our case, \(\log p(\mb{x})\) from \eqref{eq_log_marginal_b} cannot be 
evaluated, because the KL divergence from variational distribution \(q\) to the 
true posterior \(p\) cannot be computed; as the true posterior is intractable 
\eqref{eq_ivec_post_b}. We can only compute \(\mathcal{L}(q)\), which is a 
lower bound on \(\log p(x)\); thus the resulting perplexity values act as upper 
bounds. This is true for NVDM~\cite{NVI:2016} or any other model in the VB 
framework where the true posterior is intractable~\cite{Bishop:2006:PRML}.
We estimated \(\mathcal{L}(q)\) from \eqref{eq_complete_doc_elbo} using \(32\) 
samples, i.e., \(R=32\), in order to compute perplexity. In~\cite{NVI:2016}, 
the authors used \(20\) samples.
\begin{figure*}[t!]
	\centering
	\begin{subfigure}[t]{0.45\textwidth}
		\includegraphics[width=\linewidth]{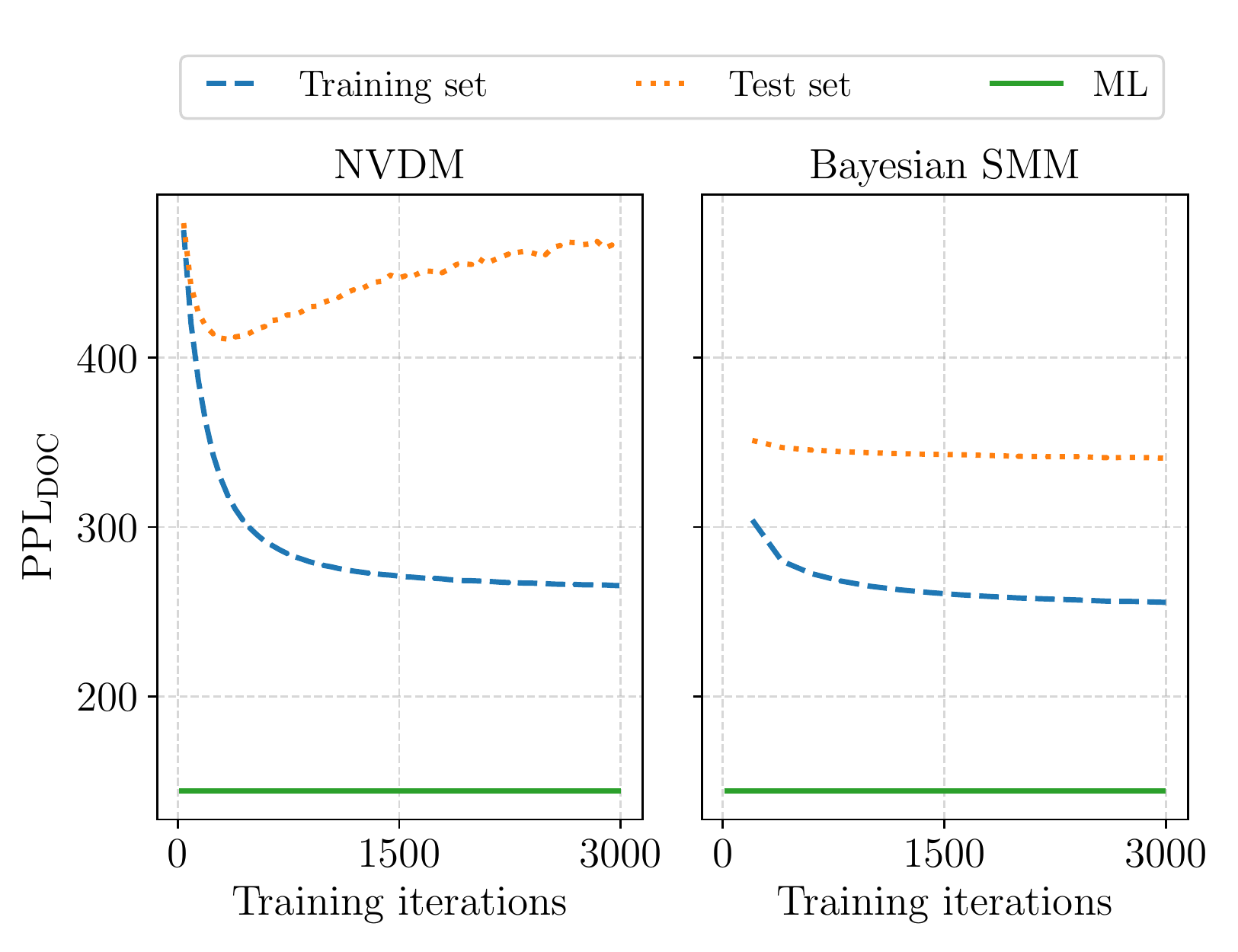}
		\caption{\label{fig:ppl_man} PPL of \textit{Fisher} test data.}
	\end{subfigure}
	\begin{subfigure}[t]{0.45\textwidth}
		\includegraphics[width=\linewidth]{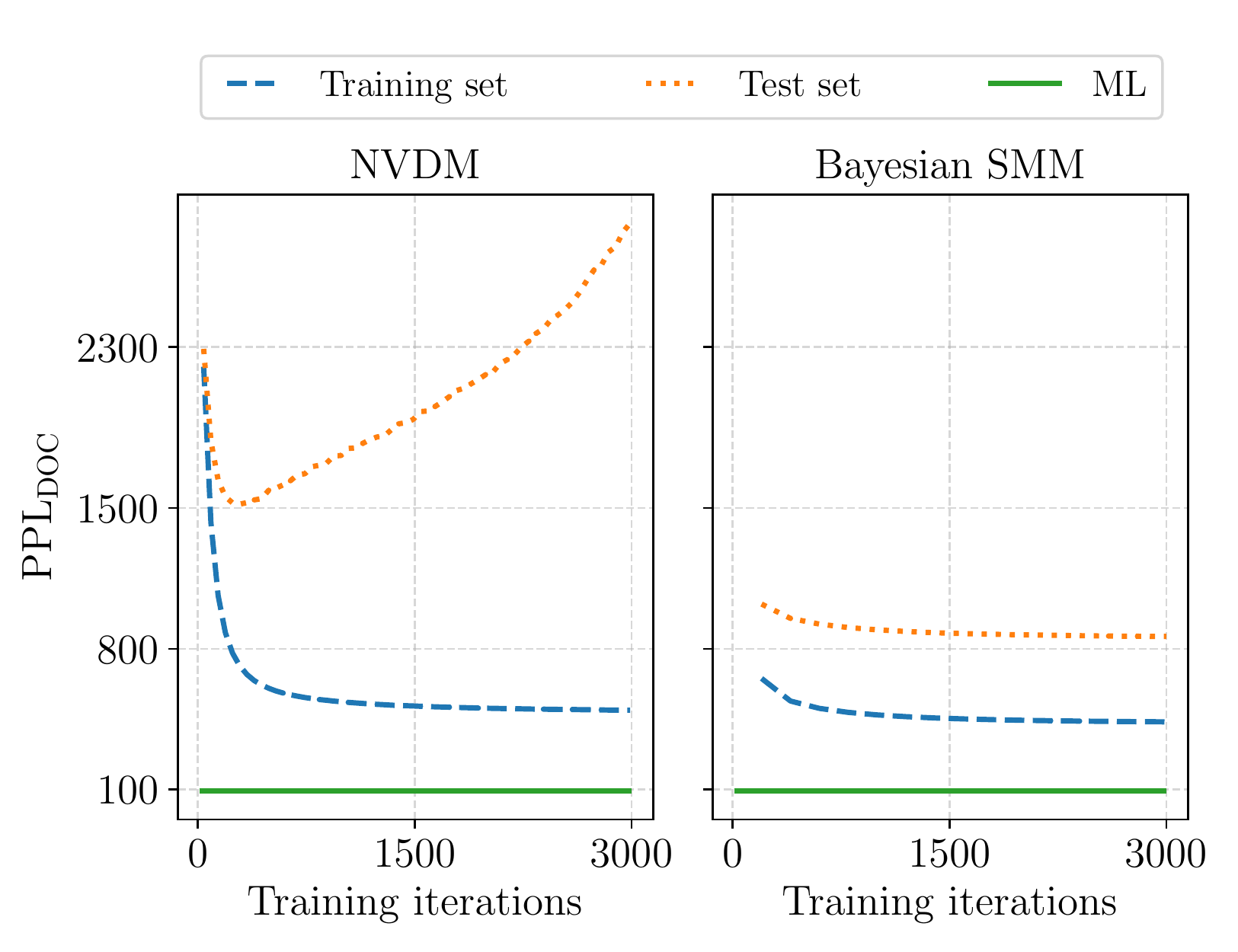}
		\caption{\label{fig:ppl_20news} PPL of \textit{20Newsgroups} test data.}
	\end{subfigure}
	\caption{Comparison of training and test data perplexities obtained using 
	Bayesian SMM and NVDM for both \textit{Fisher} and \textit{20Newsgroups} datasets. The 
	horizontal solid green line shows the test data perplexity computed using the maximum 
	likelihood (ML) probabilities estimated on the test data. The latent (embedding) dimension 
	was set to \(200\) for both the models.}
\end{figure*}

We present the comparison of \textit{20Newsgroups} test data perplexities 
obtained using Bayesian SMM and NVDM in Table~\ref{tab:ppl}. It shows the 
perplexities of \textit{20Newsgroups} corpus under full and a limited 
vocabulary of 2000 words~\cite{NVI:2016}. We also show the perplexity computed 
using the maximum likelihood probabilities estimated on the test data. It acts 
as the lower bound on the test perplexities. NVDM was shown~\cite{NVI:2016} to 
achieve superior perplexity scores when compared to LDA, 
docNADE~\cite{Hugo:2012:DocNADE}, Deep Auto Regressive Neural Network 
models~\cite{Mnih:2014:NVI}. To the best of our knowledge, our model achieves 
state-of-the-art perplexity scores on \textit{20Newsgroups} corpus under 
limited and full vocabulary conditions.


In further investigation, we trained both Bayesian SMM and NVDM until 
convergence. At regular checkpoints during the training, we froze the model, 
extracted the embeddings for both training and test data, and computed the 
perplexities; shown in Figures~\ref{fig:ppl_man} and \ref{fig:ppl_20news}. We 
can observe that both the Bayesian SMM and NVDM fit the training data equally 
well (low perplexities). However, in the case of NVDM, the perplexity of test 
data increases after certain number of iterations; suggesting that NVDM fails 
to generalize and over-fits on the training data. In the case of Bayesian SMM, 
the perplexity of the test data decreases and remains stable, illustrating the 
robustness of our model. 

%
%
\subsection{Early stopping mechanism for topic ID systems}
\label{ssec:ease_of_training}
\begin{figure*}[t!]
  \centering
  \includegraphics[width=0.9\linewidth]{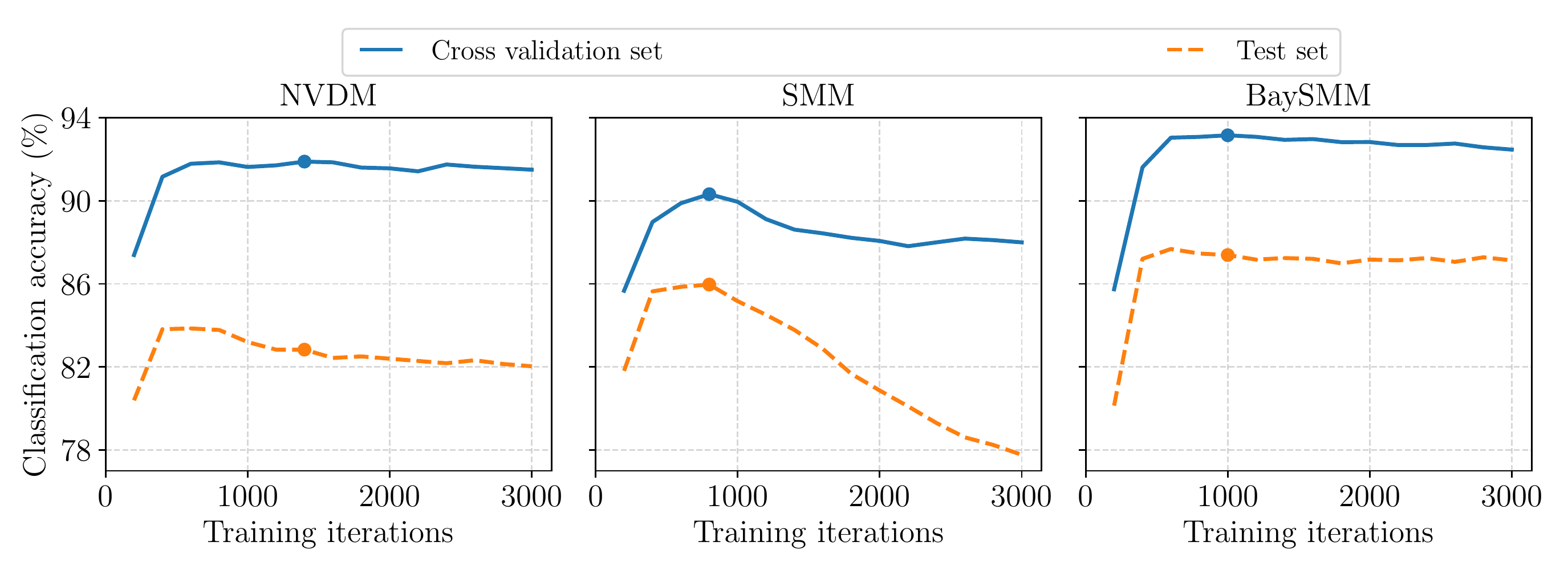}
  \caption{\label{fig:acc_conv} Performance of topic ID systems on     
  \textit{Fisher} data at various checkpoints during model training. The 
  circular dot (\(\bullet\)) represents the best cross-validation score and the 
  corresponding test score obtained using the early stopping mechanism (ESM). 
  The embedding dimension was set to 100 for all the models.} 
\end{figure*}
The embeddings extracted from a model trained purely in an unsupervised fashion 
does not necessarily yield optimum results when used in a supervised 
scenario. As discussed earlier in Sections~\ref{ssec:proposed_sys}, 
and~\ref{ssec:baselines}, an early stopping mechanism (ESM) during the training 
of an unsupervised model (eg: NVDM, SMM, and Bayesian SMM) is required to get 
optimal performance from the subsequent topic ID system. The following 
experiment illustrates the idea of ESM:

We trained SMM, Bayesian SMM and NVDM on \textit{Fisher} data until 
convergence. At regular checkpoints during the training, we froze the model, 
extracted the embeddings for both training and test data. We chose GLC for SMM, 
GLCU for NVDM, and Bayesian SMM as topic ID classifiers. We then evaluated 
the topic ID accuracy on the cross-validation\footnote{5-fold cross-validation 
on training set.} and test sets.  Fig.~\ref{fig:acc_conv} shows the topic ID 
accuracy on cross-validation and test sets obtained at regular checkpoints for 
all the three models. The circular dot (\(\bullet\)) represents the best 
cross-validation score and the corresponding test score that is obtained by 
employing ESM. In case of (non-Bayesian) SMM, the test 
accuracy drops significantly after certain number of iterations; suggesting the 
strong need of ESM. The cross-validation accuracies of NVDM and 
Bayesian SMM are similar and remain consistent over the iterations. However, 
the test 
accuracy of NVDM is much lower than that of Bayesian SMM and also decreases 
over the iterations. On the other hand, the test accuracy of Bayesian SMM 
increases and stays consistent. It shows the robustness of our proposed model, 
which in addition, does not require any ESM. In all the further 
topic ID experiments, we report classification results for Bayesian SMM without 
ESM; while the results for SMM, and NVDM are with ESM.

\subsection{Topic ID results}
\label{ssec:topic_id}
%
%
%
%
\begin{table*}[!t]
 \begin{center}
 \caption{\label{tab:res}Comparison of results on \textit{Fisher} test sets, 
	from earlier published works, our baselines and proposed systems. \(\star\) 
	indicates a pure discriminative model.}
 \begin{tabular}{llcrrrr} \toprule
  \bf{Systems} & \bf{Model} & \bf{Classifier} & \bf{Accuracy (\%)} & \bf{CE} & 
  \bf{Accuracy (\%)} & \bf{CE} \\ \midrule
 & & & \multicolumn{2}{c}{\bf{Manual transcriptions}} & 
 \multicolumn{2}{c}{\bf{Automatic transcriptions}}\\ \midrule
 \multirow{2}{*}{Prior works} & BoW~\cite{Hazen:2007:ASRU} & NB & 87.61 &-&-&-\\
  &TF-IDF ~\cite{May:2015:mivec} & LR   & 86.41       &  - & - & -     \\
	\midrule
  \multirow{7}{*}{Our Baseline} & TF-IDF & LR & 86.59 & 0.93 & 86.77 & 
  \textbf{0.94} \\
 &ULMFiT \(\star\)   & MLP  & 86.41 & \textbf{0.50} & 86.08 & \textbf{0.50} 
		\\ \cmidrule{2-7}
 &\(\ell_1\) SMM      & LR   & 86.81 & 0.91 & 87.02 & 1.09 \\
 &\(\ell_1\) SMM      & GLC  & 85.17 & 1.64 & 85.53 & 1.54 \\ \cmidrule{2-7}
 &NVDM                       & LR    & 81.16 & 0.94 & 83.67 & 1.15 \\
 &NVDM                       & GLC   & 84.47 & 1.25 & 84.15 & 1.22 \\
 &NVDM                       & GLCU  & 83.96 & 0.93 & 83.01 & 0.97 \\ \midrule
 \multirow{3}{*}{Proposed} &Bayesian SMM& LR& \textbf{89.91} &  \textbf{0.89} 
 & \textbf{88.23} & 0.95\\
 &	Bayesian SMM  & GLC & \textbf{89.47} & 1.05 & \textbf{87.23} & 1.46 \\
 &	Bayesian SMM & GLCU   & \textbf{89.54} & \textbf{0.68} & 		
 \textbf{87.54} & \textbf{0.77} \\
 \bottomrule
 \end{tabular}
 \end{center}
\end{table*}
This section presents the topic ID results in terms of classification 
accuracy (in \%) and cross-entropy (CE) on the test sets. Cross-entropy gives
a notion of how confident the classifier is about its prediction. A well 
calibrated classifier tends to have lower cross-entropy.

Table~\ref{tab:res} presents the classification results on \textit{Fisher} 
speech corpora with manual and automatic transcriptions, where the first two 
rows are the results from earlier published works.
Hazen~\cite{Hazen:2007:ASRU}, used  discriminative vocabulary selection
followed by a na\"ive Bayes (NB) classifier. Having a limited (small) 
vocabulary is the major drawback of this approach. Although we have used the 
same training and test splits, May~\cite{May:2015:mivec} had slightly larger 
vocabulary than ours, and their best system is similar to our baseline TF-IDF
based system. The remaining rows in Table~\ref{tab:res} show our baselines 
and proposed systems. We can see that our proposed systems achieve 
consistently better accuracies; notably, GLCU which exploits the uncertainty in 
document embeddings has much lower cross-entropy than its counter part, GLC. To 
the best of our knowledge, the proposed systems achieve the best classification 
results on \textit{Fisher} corpora with the current set-up, i.e., treating each 
side of the conversation as an independent document. It can be observed ULMFiT 
has the lowest cross-entropy among all the systems.

Table~\ref{tab:res2} presents classification results on 
\textit{20Newsgroups} dataset. The first three rows give the results as 
reported in earlier works. Pappagari et al.~\cite{Raghu:2018:CNN}, proposed a 
CNN-based discriminative model trained to jointly optimize categorical 
cross-entropy loss for classification task along with binary cross-entropy 
for verification task. Sparse composite document vector 
(SCDV)~\cite{Mekala:2017:SCDV} exploits pre-trained word embeddings to obtain 
sparse document embeddings, whereas neural tensor skip-gram model
(NTSG)~\cite{Liu:2015:NTSG} extends the idea of a skip-gram model for
obtaining document embeddings. The authors in (SCDV)~\cite{Mekala:2017:SCDV} 
have shown superior classification results as compared to paragraph vector, 
LDA, NTSG, and other systems. The next rows in Table~\ref{tab:res2} present our 
baselines and proposed systems. We see that the topic ID systems based on 
Bayesian SMM and logistic regression is better than all the other models, 
except for the purely discriminative CNN model. We can also see that all the 
topic ID systems based on Bayesian SMM are consistently better than variational 
auto encoder inspired NVDM, and (non-Bayesian) SMM. 

The advantages of the proposed Bayesian SMM are summarized as follows: (a) the 
document embeddings are Gaussian distributed which enables to train simple  
generative classifiers like GLC, or GLCU; that can extended to newer classes 
easily, (b) although the Bayesian is trained in an unsupervised fashion, it 
does not require any early stopping mechanism to yield optimal topic ID 
results; document embeddings extracted from a fully converged or model can be  
directly used for classification tasks without any fine-tuning.

\begin{table}[!t]
	\begin{center}
		\caption{\label{tab:res2}Comparison of results on 
		\textit{20Newsgroups} from earlier published works, our baselines and 
		proposed systems. \(\star\) indicates a pure discriminative model.}
		\begin{tabular}{llcrr} \toprule
 		\bf{Systems} & \bf{Model} & \bf{Classifier} & \bf{Accuracy (\%)} & \bf{CE} \\
		\midrule
  \multirow{3}{*}{Prior works} & CNN~\cite{Raghu:2018:CNN} \(\star\) & 
  	- & \textbf{86.12} 	& - \\
 & SCDV~\cite{Mekala:2017:SCDV} & SVM  &  \textbf{84.60} & - \\
 &	NTSG-1~\cite{Liu:2015:NTSG} & SVM  &  82.60 & -  \\ \midrule
 \multirow{7}{*}{Our Baselines} & TF-IDF   & LR  & 84.47 & \textbf{0.73}  \\
 & ULMFiT \(\star\) & MLP  & 83.06 & 0.89 \\ \cmidrule{2-5}
 &	\(\ell_1\) SMM  & LR   & 82.01 & \textbf{0.75} \\
 &	\(\ell_1\) SMM  & GLC  & 82.02 &  1.33 \\  \cmidrule{2-5}
 & NVDM             & LR   & 79.57 & 0.86 \\
 & NVDM             & GLC  & 77.60 & 1.65\\
 & NVDM             & GLCU & 76.86 & 0.88 \\	\midrule
 \multirow{3}{*}{Proposed} & Bayesian SMM & LR & \textbf{84.65} & \textbf{0.53} 
 \\
 & Bayesian SMM           & GLC    &  83.22 & 1.28 \\
 & Bayesian SMM           & GLCU   &  82.81 & 0.79 \\
	\bottomrule
 \end{tabular}
 \end{center}
\end{table}
\subsection{Uncertainty in document embeddings}
\label{ssec:uncert}
\begin{figure}[ht!]
	\centering
	\includegraphics[width=\linewidth]{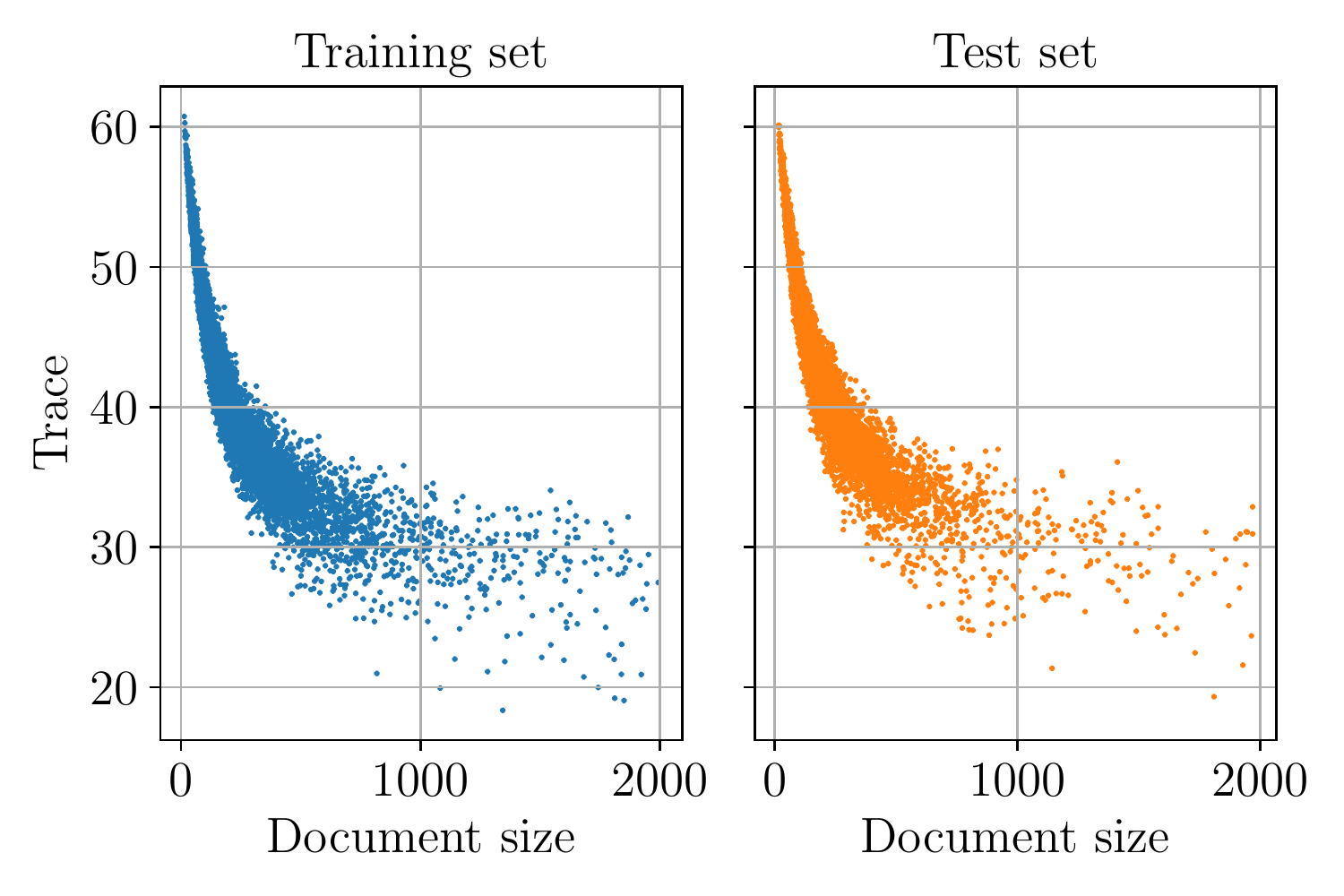}
	\caption{\label{fig:trace} Uncertainty (trace of covariance of posterior 
		distribution) captured in the document embeddings of 
		\textit{20Newsgroups} dataset.}
\end{figure}

The uncertainty captured in the posterior distribution of document embeddings
correlates strongly with size of the document. The trace of the covariance matrix of the inferred posterior distributions gives us the notion of such a correlation. Fig.~\ref{fig:trace} shows an example of uncertainty captured in
the embeddings. Here, the Bayesian SMM was trained on \textit{20Newsgroups} 
with an embedding dimension of 100. 

\section{Conclusions and future work}
\label{sec:concl}
We have presented a generative model for learning document representations 
(embeddings) and their uncertainties. Our proposed model achieved 
state-of-the-art perplexity results on the standard \textit{20Newsgroups} and 
\textit{Fisher} datasets. Next, we have shown that the proposed model is robust 
to over-fitting and unlike in SMM and NVDM, it does not require any early 
stopping mechanism for topic ID. We proposed an extension to simple Gaussian 
linear classifier that exploits the uncertainty in document embeddings and 
achieves better cross-entropy scores on the test data as compared to the simple 
GLC. Using simple linear classifiers on the 
obtained document embeddings, we achieved superior classification results on 
\text{Fisher} speech \textit{20Newsgroups} text corpora. We also addressed a 
commonly encountered problem of intractability while performing variational 
inference in mixed-logit models by using the re-parametrization trick. This 
idea can be translated in a straightforwardly for subspace \(n\)-gram model for 
learning sentence embeddings and also for learning word embeddings along with 
their uncertainties. The proposed Bayesian SMM can be extended to have 
topic-specific priors for document embeddings, which enables to encode topic 
label uncertainty 
explicitly in the document embeddings. There exists other scoring mechanisms 
that exploit the uncertainty in embeddings~\cite{Brummer:2018:GE}, which we 
plan to explore in our future works.
%

%


\appendices

\section{Gradients of Lower Bound}
%
%

The variational distribution is diagonal with the following parametrization:
\begin{equation}
q(\mb{w}) = \mathcal{N}(\mb{w} \,|\, \mb{\nu}, \mathrm{diag}(\exp\{2\mb{\varsigma}\})).
 \end{equation}
The lower bound for a single document is:
\begin{multline}
\mathcal{L}_d \approx -\frac{1}{2} \Bigg[
\lambda \, \mathrm{tr}(\mathrm{diag}(\exp\{2\mb{\varsigma}\})) 
- \log|\mathrm{diag}(\exp\{2\mb{\varsigma}\})| \\ - K \, \log \lambda +  \lambda \mb{\nu}^{\T}\mb{\nu} - K\Bigg] \\
+\sum_{i=1}^{V} x_i \, \Bigg[\,(m_i + \mb{t}_i \bs{\nu}) \\ 
\- \frac{1}{R}\sum_{r=1}^{R} \log (\sum_{j=1}^{V} \exp\{m_j + \mb{t}_j\,g(\bs{\epsilon}_{r})\}) \Bigg], \\
 \label{eq:app_obj_baysmm}
\end{multline}
where
\begin{equation}
g(\mb{\epsilon}) = \mb{\nu} + \mathrm{diag}(\exp\{\mb{\varsigma}\}) \tileps.
\end{equation}
It is convenient to have the following derivatives:
\begin{align}
\pdv{g(\veps)}{\vnu} &= \mI \label{eq:app_derv_g_nu}. \\
\pdv{(\mb{t}_i g(\bs{\epsilon}))}{\mb{\varsigma}} &= \mathrm{diag}(\mb{t}_i^{\T})\, \mathrm{diag}(\exp\{\mb{\varsigma}\})\, \mathrm{diag}(\tileps) \nonumber \\
&= \mb{t}_i^{\T} \odot \exp\{\mb{\varsigma}\} \odot \tileps. \label{eq:app_derv_g_l}
\end{align}
\subsection*{Derivatives of the parameters of variational distribution:}
Taking derivative of the objective function (\eqref{eq:app_obj_baysmm}) with 
respect to mean parameter \(\vnu\) and using \eqref{eq:app_derv_g_nu}:
\begin{align}
\pdv{\mathcal{L}_d}{\bs{\nu}} &= -\lambda \mb{\nu} + \sum_{i=1}^V x_i \Bigg[\mb{t}_i^{\T} - \nonumber \\
&{} \quad \frac{1}{R}\sum_{r=1}^{R} \sum_{k=1}^{V} \mb{t}_k^{\T} \mI \, \underbrace{\myfrac{\exp\{m_k + \mb{t}_k \, g(\bs{\epsilon}_{r})\}}{\sum_j \exp\{m_j + \mb{t}_j\, g(\bs{\epsilon}_{r})\}}}_{\theta_{kr}} \Bigg] \\
&= \Big[\sum_{i=1}^{V} x_i \mb{t}_i^{\T} - \sum_{i=1}^{V} \mb{t}_i^{\T} \, 
\frac{1}{R} \sum_{r=1}^{R} \theta_{ir}  \sum_{k=1}^{V} x_k \Big] - \lambda 
\bs{\nu} 
\end{align}
\begin{equation}
\boxed{\nabla \bs{\nu} = \Big[\sum_{i=1}^{V} \mb{t}_i^{\T} \big( x_i - 
\frac{1}{R} \sum_{r=1}^{R} \theta_{ir} \sum_{k=1}^{V} x_k \big) \Big] - \lambda 
\bs{\nu}.} \label{eq_derv_du_rp} 
\end{equation}
Taking the derivative of objective function (\eqref{eq:app_obj_baysmm}) with 
respect to \(\mb{\varsigma}\) and using \eqref{eq:app_derv_g_l}:
\begin{align}
\pdv{\mathcal{L}_d}{\mb{\varsigma}} &= -\frac{1}{2}\Big[2 \lambda \exp\{2\mb{\varsigma}\} -2\mb{I} \Big] \nonumber \\
&{} + \sum_{i=1}^V x_i \Bigg[- \frac{1}{R} \sum_{r=1}^{R} \sum_{k=1}^{V} \mb{t}_k^{\T} \bs{\epsilon}_{r}^{\T}  \underbrace{\myfrac{\exp\{m_k + \mb{t}_k g(\bs{\epsilon}_{r})\}}{\sum_j \exp\{m_j + \mb{t}_j g(\bs{\epsilon}_{r})\}}}_{\theta_{kr}}  \Bigg] \nonumber \\
&= \mb{1} - \lambda \exp\{2\mb{\varsigma}\} \nonumber \\
{}& \qquad - \Big[\sum_{i=1}^V x_i \, \frac{1}{R}\sum_{r=1}^R \sum_{k=1}^V 
\mb{t}_k^{\T} \odot \exp\{\mb{\varsigma}\} \odot \tileps_r^{\T} 
\theta_{kr}\Big] 
\end{align} 
\begin{empheq}[box=\fbox]{align}
\nabla \mb{\varsigma} &= \mb{1} - \lambda \exp\{2\mb{\varsigma}\} \nonumber \\
{}& \quad -\Bigg[ \Big(\sum_{i=1}^V x_i\Big) \frac{1}{R}\sum_{r=1}^R 
\sum_{k=1}^V  \theta_{kr} \mb{t}_k^{\T} \odot \exp\{\mb{\varsigma}\} \odot 
\tileps_r \Bigg]. \label{eq_derv_omega_rp}
\end{empheq}

\subsection*{Derivatives of the model parameters:}
Taking the derivative of complete objective \eqref{eq_complete_reg_elbo} with 
respect to a row \(\vt_k\) from matrix \(\mT\):
\begin{align}
\pdv{\mathcal{L}}{\mb{t}_k} &= \pdv{}{\mb{t}_k} \, \sum_{d=1}^{D} \sum_{i=1}^{V} x_{di} \Bigg[(m_i + \mb{t}_i \bs{\nu}_d) \nonumber \\
{}& \qquad -\frac{1}{R}\sum_{r=1}^{R} \log (\sum_{j=1}^{V} \exp\{m_j + \mb{t}_j\,g(\,\bs{\epsilon}_{r}\,) \}) \Bigg] \nonumber \\
{}& \qquad - \omega \sum_{i=1}^{V} \lvert \lvert \mb{t}_i \rvert \rvert_1 \\
&= \sum_{d=1}^{D} \Bigg[ x_{dk} \bs{\nu}_d^{\T} \nonumber \\
{}& \qquad - \sum_{i=1}^{V} x_{di}  \frac{1}{R}\sum_{r=1}^{R} g(\bs{\epsilon}_{dr})^{\T}  \underbrace{\myfrac{\exp\{m_i + \mb{t}_k \, g(\bs{\epsilon}_{dr})\}}{\sum_j \exp\{m_j + \mb{t}_j g(\bs{\epsilon}_{dr})\}} }_{\theta_{dkr}} \Bigg] \nonumber \\
&{} \qquad - \omega \, \mathrm{sign}(\mb{t}_k) \\
&= \sum_{d=1}^{D} \Bigg[ x_{dk} \bs{\nu}_d^{\T} - \sum_{i=1}^{V} x_{di}  \frac{1}{R}\sum_{r=1}^{R}  g(\bs{\epsilon}_{dr})^{\T} \theta_{dkr} \Bigg] \nonumber \\
{}& \qquad - \omega \, \mathrm{sign}(\mb{t}_k) 
\end{align}
\begin{empheq}[box=\fbox]{align}
\gt_k &= \sum_{d=1}^{D} \Bigg[ x_{dk} \bs{\nu}_d^{\T} - \Big[ \Big(\sum_{i=1}^{V} x_{di}\Big)  \frac{1}{R}\sum_{r=1}^{R} \theta_{dkr} g(\bs{\epsilon}_{dr})^{\T} \Big] \Bigg] \nonumber \\
{}& \qquad - \omega \, \mathrm{sign}(\mb{t}_k). \label{eq_derv_t_rp}
\end{empheq}

\section{EM algorithm for GLCU}

\subsection*{\textsc{E-step}:}
 Obtaining the posterior distribution of latent variable \(p(\vy_d \,|\, 
 \vnu_d, \Theta)\). Using the results from~\cite{cookbook} (p. 41,  (358)):
\begin{align*}
\label{app_eq_e_step_glcu}
\log p(\vy_d \,&|\, \vnu_d, \vh_d, \Theta) \\
{}&= \log p(\bs{\nu}_d \mid \mb{y}_d,  \mb{h}_d) 
+ \log p(\mb{y}_d) - \log p(\bs{\nu}_d)  \\
&= \log \mathcal{N}(\bs{\nu}_d \mid \bs{\mu}_{d} + \mb{y}_d, \mb{D}^{\I}) \\
&{} \quad + \log \mathcal{N}(\mb{y}_d \mid \mb{0}, \bs{\Gamma}_d^{\I}) + \mathrm{const}  \\
&=-\frac{1}{2}(\bs{\nu}_d - (\bs{\mu}_{d} + \mb{y}_d))^{\T} \mb{D} (\bs{\nu}_d - (\bs{\mu}_{d} + \mb{y}_d))  \\
&{} \quad - \frac{1}{2} \mb{y}_d^{\T} \bs{\Gamma}_d \mb{y}_d + \mathrm{const}  \\
&=-\frac{1}{2} (\mb{y}_d - (\bs{\nu}_d - \bs{\mu}_{d}))^{\T} \mb{D} (\mb{y}_d - (\bs{\nu}_d - \bs{\mu}_{d})) \\
&{} \quad - \frac{1}{2} \mb{y}_d^{\T} \bs{\Gamma}_d \mb{y}_d  + \mathrm{const} \\
 &= \mathcal{N}(\mb{y}_d \mid \mb{u}_d, \mb{V}_d^{\I}) \\
 \intertext{where \(\vu_d\) is simplified as:}
\mb{u}_d &= (\mb{D} + \bs{\Gamma}_d)^{\I}(\mb{D} (\bs{\nu}_d - \bs{\mu}_{d}) + \bs{\Gamma}_d \mb{0}) \\
&= [\mb{D}^{\I} (\mb{D} + \bs{\Gamma}_d)]^{\I} (\bs{\nu}_d - \bs{\mu}_{d})
\end{align*}
\noindent resulting in:
\begin{empheq}[box=\fbox]{align}
\mb{u}_d &= [\mb{I} + \mb{D}^{\I} \bs{\Gamma}_d]^{\I} (\bs{\nu}_d - \bs{\mu}_{d}) \\
\mb{V}_d &= \mb{D} + \bs{\Gamma}_d \label{eq_sum_of_sqr} 
\end{empheq}
\subsection*{\textsc{M-step}:}
Maximizing the auxiliary function
\begin{align}
\Theta^{\textrm{new}} &= \argmax \limits_{\Theta}\, \mathcal{Q}(\Theta, \Theta^{\textrm{old}}) \\
q(\mb{y}) &= p(\mb{y} \mid \mb{w}, \Theta^{\textrm{old}}).
\end{align}

\noindent Using the results from~\cite{cookbook}[p. 43,  (378)], the auxiliary 
function \(\mathcal{Q}(\Theta, \Theta^{\textrm{old}})\) is computed as:
\begin{align*}
\label{eq_m_step_aux}
&{}\mathcal{Q}(\Theta, \Theta^{\textrm{old}}) \\
&= \mathbb{E}_q [\sum_{d=1}^{D} \log p(\bs{\nu}_d, \mb{y}_d)] \\
&= \sum_{d=1}^{D} \mathbb{E}_q[\log p(\bs{\nu}_d \mid \mb{y}_d) ] + \mathbb{E}_q[\log p(\mb{y}_d)]  \\
&= \sum_{d=1}^D \mathbb{E}_q[\log \mathcal{N}(\bs{\nu}_d \mid \bs{\mu}_{d} + \mb{y}_d, \mb{D}^{\I})] + \mathrm{const} \\
&= \frac{D}{2} \log|\mb{D}| - \frac{1}{2}\sum_{d=1}^{D} \Big[\mathbb{E}_q[(\bs{\nu}_d - (\bs{\mu}_{d} + \mb{y}_d))^{\T} \mb{D} \\
&{} \qquad \qquad (\bs{\nu}_d - (\bs{\mu}_{d} + \mb{y}_d))] \Big] + \mathrm{const}  \\
&= \frac{D}{2}\log |\mb{D}| - \frac{1}{2} \sum_{d=1}^{D} \Big[\mathrm{tr}(\mb{D}\mb{V}_d^{\I})\\
&{} \quad + (\mb{u}_d - (\bs{\nu}_d - \bs{\mu}_{d}))^{\T} \mb{D} (\mb{u}_d - (\bs{\nu}_d - \bs{\mu}_{d})) \Big]
\end{align*}
Maximizing the auxiliary function \(\mathcal{Q}\) with respect to model parameters \(\Theta = \{\mb{M}, \mb{D}\}\)
\begin{align*}
\intertext{Taking derivative with respect to each column \(\vmu_{\ell}\) in \(\mM\) and equating it to zero:}
\pdv{\mathcal{Q}}{\bs{\mu}_{\ell}} &= -\frac{1}{2} \pdv{}{\bs{\mu}_{\ell}} \sum_{d \in \mathcal{I}_{\ell}}  \Big[ (\mb{u}_d - (\bs{\nu}_d - \bs{\mu}_{\ell}))^{\T} \mb{D} (\mb{u}_d - (\bs{\nu}_d - \bs{\mu}_{l}))  \Big] \numberthis \\
&= -\frac{1}{2} \sum_{d \in \mathcal{I}_{\ell}} \, 2 \mb{D} \, \big(\bs{\mu}_{\ell} - (\bs{\nu}_d - \mb{u}_d)\big) \\
&= -\mb{D} \, \Big( \sum_{n \in \mathcal{I}_{\ell}} \, \bs{\mu}_{\ell} - \sum_{n \in \mathcal{I}_{\ell}} \, (\bs{\nu}_d - \mb{u}_d) \Big) 
\end{align*}
\begin{empheq}[box=\fbox]{equation}
\bs{\mu}_{\ell} = \frac{1}{|\mathcal{I}_{\ell}|} \sum_{n \in \mathcal{I}_{\ell}} (\bs{\nu}_d - \mb{u}_d)\label{app_eq_mu_l_update} 
\end{empheq}
Taking derivative with respect to shared precision matrix \(\mD\) and equating it to zero:
\begin{align}
\pdv{\mathcal{Q}}{\mb{D}} &= \frac{D}{2} \mb{D}^{\I} - \frac{1}{2} \Big(\sum_{d=1}^{D} 
\mb{V}_d^{\I}\Big)^{\T} \nonumber \\
&-\frac{1}{2}\Big(\sum_{d=1}^{D} (\mb{u}_d - (\bs{\nu}_d - \bs{\mu}_{d})) (\mb{u}_d - (\bs{\nu}_d - \bs{\mu}_{nl}))^{\T}\Big)^{\T}
\end{align}
\begin{empheq}[box=\fbox]{align}
\mb{D}^{\I} &= \frac{1}{D} \Big[\sum_{d=1}^{D} \mb{V}_d^{\I} \nonumber \\
&{} \quad + \sum_{d=1}^{D} (\mb{u}_d - (\bs{\nu}_d - \bs{\mu}_{nl})) (\mb{u}_d - (\bs{\nu}_d - \bs{\mu}_{d}))^{\T}\Big]
\end{empheq}


\ifCLASSOPTIONcaptionsoff
  \newpage
\fi




\bibliographystyle{IEEEtran}
\bibliography{refs}
\end{document}